\documentclass{article}

\usepackage{microtype}
\usepackage{graphicx}
\usepackage{subcaption}
\usepackage{booktabs} 

\usepackage{hyperref}
\usepackage{booktabs}
\usepackage{tcolorbox}
\usepackage{pifont}
\usepackage{algorithm}
\usepackage{algpseudocode}

\newcommand{\cmark}{\ding{51}} 
\newcommand{\xmark}{\ding{55}} 

\usepackage[preprint]{icml2026}

\usepackage{amsmath}
\usepackage{amssymb}
\usepackage{mathtools}
\usepackage{amsthm}
\usepackage{amsmath,amssymb}
\usepackage{booktabs}
\usepackage{tabularx}
\usepackage{multirow}
\usepackage{float}

\usepackage{array}
\newcolumntype{Y}{>{\raggedright\arraybackslash}X}

\usepackage[capitalize,noabbrev]{cleveref}

\theoremstyle{plain}

\theoremstyle{definition}

\theoremstyle{remark}

\usepackage[textsize=tiny]{todonotes}

\icmltitlerunning{TDGNet: Hallucination Detection in Diffusion Language Models via Temporal Dynamic Graphs}

\begin{document}

\twocolumn[
  \icmltitle{TDGNet: Hallucination Detection in \\ Diffusion Language Models via Temporal Dynamic Graphs}



  \icmlsetsymbol{equal}{*}

  \begin{icmlauthorlist}
    \icmlauthor{Arshia Hemmat}{comp}
    \icmlauthor{Philip Torr}{eng}
    \icmlauthor{Yongqiang Chen}{cmu,mbzui}
    \icmlauthor{Junchi Yu}{eng}

  \end{icmlauthorlist}

  \icmlaffiliation{comp}{Department of Computer Science, University of Oxford}
  \icmlaffiliation{eng}{Department of Engineering Science, University of Oxford}
  \icmlaffiliation{cmu}{Carnegie Mellon University}
  \icmlaffiliation{mbzui}{Mohamed bin Zayed University of Artificial Intelligence}

  \icmlcorrespondingauthor{Junchi Yu}{junchi.yu@eng.ox.ac.uk}

  \icmlkeywords{Large Language Models, Hallucination Detection, Graph Neural Network, ICML}

  \vskip 0.3in
]



\printAffiliationsAndNotice{}  

\begin{abstract}
Diffusion language models (D-LLMs) offer parallel denoising and bidirectional context, but hallucination detection for D-LLMs remains underexplored. Prior detectors developed for auto-regressive LLMs typically rely on single-pass cues and do not directly transfer to diffusion generation, where factuality evidence is distributed across the denoising trajectory and may appear, drift, or be self-corrected over time. We introduce TDGNet, a temporal dynamic graph framework that formulates hallucination detection as learning over evolving token-level attention graphs. At each denoising step, we sparsify the attention graph and update per-token memories via message passing, then apply temporal attention to aggregate trajectory-wide evidence for final prediction. Experiments on LLaDA-8B and Dream-7B across QA benchmarks show consistent AUROC improvements over output-based, latent-based, and static-graph baselines, with single-pass inference and modest overhead. These results highlight the importance of temporal reasoning on attention graphs for robust hallucination detection in diffusion language models.
\end{abstract}

\section{Introduction}
\begin{figure*}
    \centering
    \includegraphics[width=0.95\textwidth]{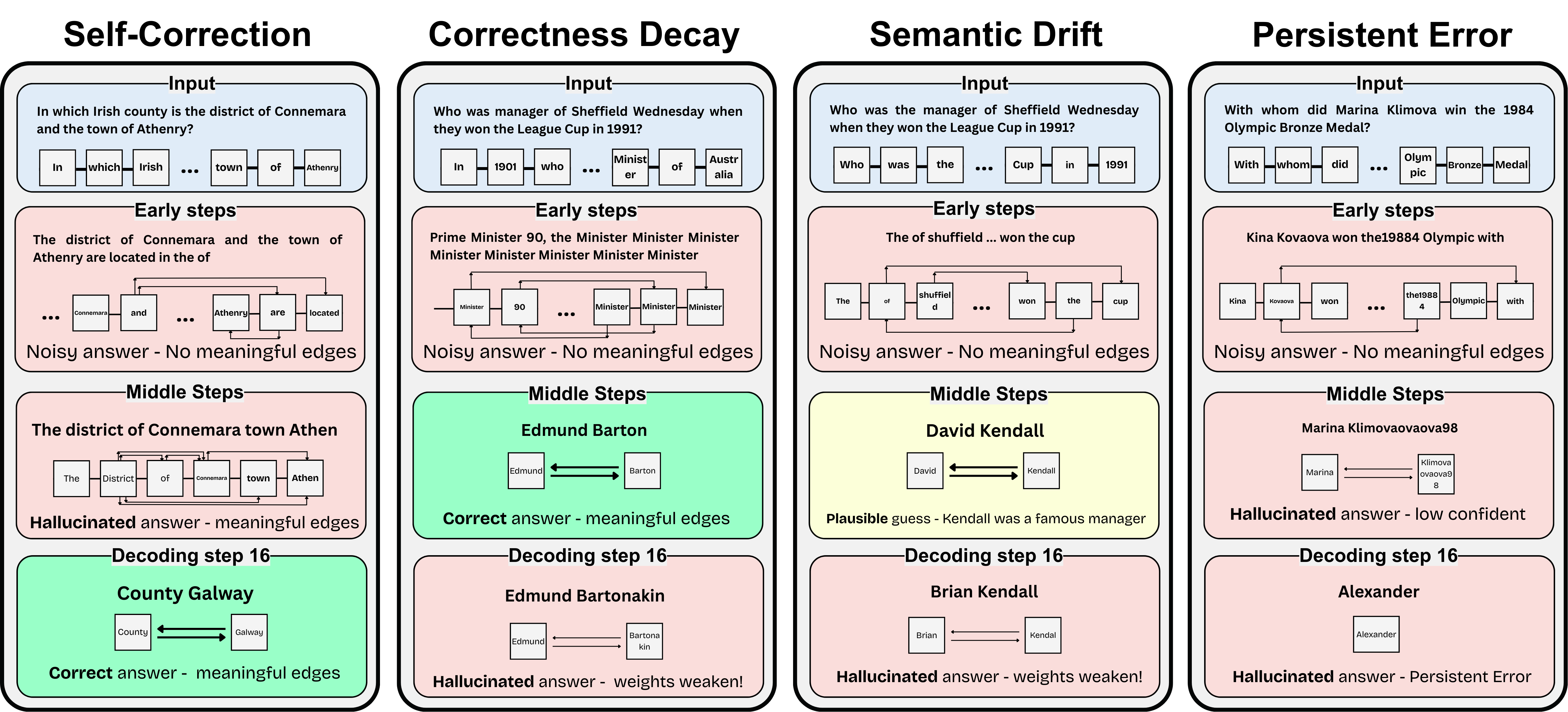}
    \caption{\textbf{Why Temporal Modeling is Essential.} We visualize four diffusion dynamics that confound static baselines: \textbf{Self-Correction} (early noise resolves into fact); \textbf{Correctness Decay} and \textbf{Semantic Drift} (where factual or plausible states degrade into hallucinations); and \textbf{Persistent Error} (where errors lock in early). TDGNet aggregates structural signals across the trajectory sequences to distinguish these evolving patterns, avoiding the pitfalls of single-snapshot detectors.}
    \label{fig:concept_overview}
\end{figure*}

The auto-regressive large language models (AR-LLMs) have become the \textit{de facto} paradigm across a wide range of applications, from natural language understanding ~\citep{brown2020language, ouyang2022training, openai2023gpt4}
to general-purpose task execution~\citep{vaswani2017attention, brown2020language, touvron2023llama}.
Recently, diffusion-based large language models (D-LLMs)~\citep{nie2025lldm, ye2025dream7b, sahoo2024simple} have emerged as a compelling alternative to their auto-regressive counterparts.
In contrast to AR-LLMs, which generate text sequentially via next-token prediction, D-LLMs leverage bi-directional attention and iteratively denoise the entire output sequence in generation.
This formulation enables more flexible generation dynamics ~\citep{li2022diffusion, mireshghallah2022mix} and can significantly improve inference efficiency through parallel decoding mechanisms~\citep{zhang2025surveyparallel, ghazvininejad2019mask}.
Recent representative models, such as LLaDA-8B ~\citep{nie2025lldm} and Dream-7B ~\citep{ye2025dream7b}, have demonstrated that D-LLMs can be successfully scaled to the billion-parameter scale with comparable performance to popular AR-LLMs of similar size across a wide range of diverse tasks ~\citep{nie2025lldm,ye2025dream7b}. 

As the capabilities of D-LLMs increase significantly, it becomes increasingly critical to ensure the generated content from D-LLMs is factually correct. 
However, the hallucination detection problem remains substantially less explored than that in AR-LLMs. 
Existing methods designed for AR models are fundamentally ill-suited for the diffusion paradigm due to two architectural mismatches. 
First, the \emph{probabilistic intractability} of diffusion generation prevents the computation of reliable single-pass confidence scores of the responses without expensive trajectory marginalization~\citep{kuhn2023semantic,sahoo2024simple}. 
Second, D-LLMs exhibit \emph{temporal oscillation}, where predictions fluctuate between factual and hallucinated states during the iterative denoising process~\citep{wang2025time}. Such a phenomenon demonstrates that hallucinations manifest as dynamic events in the generation process of D-LLMs, such as self-correction, late-stage drift, or persistence, rather than static errors in the model responses.
Thus, effective hallucination detection in D-LLMs
requires modeling the temporal evolution of the generation trajectory rather than inspecting isolated snapshots of the output~\citep{chang2025tracedet}.

Building on the observation that hallucinations in D-LLMs unfold over the denoising trajectory (Section~\ref{sec:fate_analysis}), we focus on how such dynamics should be represented for effective detection.
While one may attempt to model the generation process as a sequence of intermediate token states, this abstraction fails to capture the underlying structure in the intermediate contexts.
Unlike AR-LLMs, D-LLMs rely on bi-directional attention to iteratively refine global context, which introduces evolving correlations among tokens during the generation process.
This evolving pattern of token correlations suggests a representation that explicitly models both relational structure and temporal evolution.

In this work, we propose TDGNet, a temporal graph–based framework for hallucination detection in diffusion-based LLMs that explicitly models the evolving relational structure of the denoising process. 
The motivation is that hallucinations in D-LLMs arise from changes in token interactions over time, which cannot be reliably captured by sequence-based or snapshot-based representations.
Specifically, TDGNet formulates hallucination detection as a temporal graph learning problem, where tokens are treated as nodes and their interactions evolve across denoising steps.
TDGNet maintains a persistent memory state ~\citep{rossi2020tgn} for each token that is updated through temporal message passing. 
This design enables the TDGNet to aggregate subtle signs of hallucination from the token dependency across the denoising generation process of D-LLMs.
Moreover, TDGNet naturally supports both sequence-level and token-level hallucination detection, enabling identification of hallucination at multiple granularities.

\begin{figure*}
    \centering
    \includegraphics[width=1\linewidth]{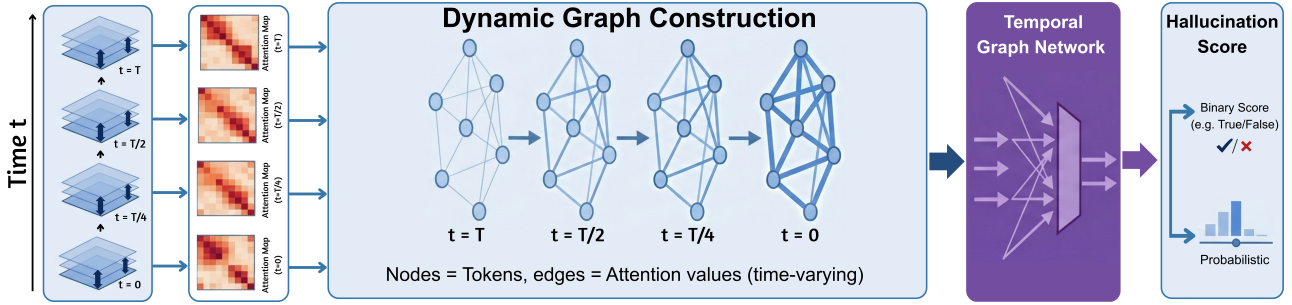}
    \caption{\textbf{Overview of the TDGNet framework.} The model extracts attention maps across diffusion denoising steps to construct a temporal dynamic graph, which is then processed by a Temporal Graph Network to predict hallucination probability.}
    \label{fig:placeholder}
\end{figure*}

We evaluate our method on two open-source D-LLMs, LLaDA-8B and Dream-7B ~\citep{nie2025lldm, ye2025dream7b}, across multiple-choice, open-ended, and contextual QA benchmarks, using AUROC as the primary metric for hallucination detection. Our temporal graph detector consistently outperforms strong output-based, latent-based, and attention/graph-based baselines, on \emph{all} datasets and on \emph{both} models, and remains robust under different denoising schedules and
graph-construction choices (Tables~\ref{tab:baseline_vs_graph_combined}, \ref{tab:graph_variations_combined}, and~\ref{tab:transfer_method_grouped}). Our main contributions are:

    
    

\begin{itemize}
    \item We analyze hallucination dynamics in D-LLMs and show that factuality signals are weak at any single denoising step and instead accumulate across the denoising trajectory (Section~\ref{sec:fate_analysis}).
    
    \item We propose \textbf{TDGNet}, a temporal graph learning framework that tracks token interactions with persistent memories, enabling both response-level detection and token-level localization in a unified model (Section~\ref{sec:method}).
    
    \item Across LLaDA-8B and Dream-7B on QA benchmarks, TDGNet achieves consistent AUROC improvements over output-based and latent-based baselines, supported by ablations and cross-model evaluations that both attention-graph structure and temporal modeling are essential for strong performance. (Section~\ref{sec:res}).
\end{itemize}

\label{sec:intro}

\section{Related Work}
Hallucination Detection is a central problem for safe and truthful deployment of LLM-based systems.
Most existing methods are designed for AR-LLMs and are commonly grouped into:
(i) {Output-based} approaches, which use output-derived signals such as semantic entropy and lexical similarity, as well as other uncertainty measures and self-consistency signals
\citep{chang2025tracedet,kuhn2023semantic,lin2023generating,kossen2024semantic,manakul2023selfcheckgpt};
and (ii) {Latent-based} approaches, which probe internal states (hidden representations) to directly separate truthful vs hallucinated generations
\citep{chang2025tracedet,azaria2023internal,chen2024inside,burns2022ccs,park2025tsv,orgad2024llmsknow,du2024haloscope,su2024realtime}.
In addition, retrieval-augmented or tool-based factuality checking methods can be effective, but they rely on external knowledge sources and extra pipelines, which is not always aligned with a pure internal-signal detection objective \citep{min2023factscore,chern2023factool}.
 
Diffusion-style language models generate text by iteratively denoising a corrupted sequence, enabling refinement-based decoding rather than strict left-to-right next-token prediction \citep{zhang2025surveyparallel,lou2024discrete}.
Recent discrete diffusion objectives and reparameterized decoding schemes have improved the practicality of diffusion for language modeling and infilling \citep{lou2024discrete,zheng2023reparam}.
More broadly, diffusion decoding is increasingly framed as a key branch of parallel text generation with an explicit compute, quality trade-off \citep{zhang2025surveyparallel}. Diffusion Large Language Model extends diffusion models to text generation \citep{li2022diffusion,sahoo2024simple}. However, hallucination detection for D-LLMs is still underexplored, and existing AR-LLM detectors face practical mismatches because D-LLMs are generated through a \emph{multi-step denoising process} and may not expose the same token-logit signals as AR decoding \citep{chang2025tracedet}.

Graph neural networks (GNNs) provide a general framework for learning on relational structures by iteratively passing and aggregating messages along edges \citep{gilmer2017mpnn,battaglia2018relational}.
Popular architectures include graph convolutions and attention-based layers \citep{kipf2017gcn,hamilton2017graphsage,velickovic2018gat,xu2019gin}.
When interactions evolve, temporal/dynamic GNNs incorporate time encodings and persistent node memories to model sequences of graph snapshots or continuous-time event streams \citep{pareja2020evolvegcn,sankar2020dysat,trivedi2019dyrep,kumar2019jodie,xu2020tgat,rossi2020tgn}.
In transformer-based models, self-attention naturally induces a weighted directed token graph whose edge weights and node states change across layers and iterative decoding steps, making temporal graph learning a convenient abstraction for modeling and aggregating these evolving dependency structures \citep{vaswani2017attention}.

\label{sec:dataset}

\section{Method}
\label{sec:method}
\begin{table*}[t]
\centering
\small
\setlength{\tabcolsep}{6pt}
\renewcommand{\arraystretch}{1.08}
\caption{\textbf{Hallucination Detection Performance (AUROC) on LLaDA and Dream-7B.} We compare TDGNet against output-, latent-, and graph-based baselines across four QA benchmarks. TDGNet consistently achieves state-of-the-art performance, outperforming strong baselines (e.g., TSV) by up to 8.5\%. These results confirm that modeling the denoising trajectory yields superior robustness compared to static output statistics or isolated snapshots. (SS: Single Sampling).}
\label{tab:baseline_vs_graph_combined}
\begin{tabular}{l l c r r r r r}
\toprule
\textbf{Model} & \textbf{Metric} & \textbf{SS} & \textbf{Math} & \textbf{CSQA} & \textbf{HotpotQA} & \textbf{TriviaQA} & \textbf{Ave} \\
\midrule
\multirow{8}{*}{\textbf{LLaDA-8B-Instruct}}
   & \multicolumn{7}{c}{\textbf{Output-Based}} \\
\cmidrule(l){2-8}
   & Semantic Entropy   & \xmark & 0.68 & \underline{0.64} & \underline{0.61} & \underline{0.66} & \underline{0.65} \\
   & Lexical Similarity & \xmark & 0.51 & 0.53 & 0.54 & 0.54 & 0.53 \\
   & LN-Entropy         & \xmark & 0.70 & 0.59 & 0.55 & 0.55 & 0.60 \\
   & Perplexity         & \xmark & 0.67 & 0.60 & 0.51 & 0.54 & 0.58 \\
\cmidrule(l){2-8}
   & \multicolumn{7}{c}{\textbf{Latent-Based}} \\
\cmidrule(l){2-8}
   & EigenScore         & \xmark & 0.56 & 0.54 & 0.56 & 0.59 & 0.56 \\
   & TSV                & \cmark & \underline{0.72} & 0.61 & 0.55 & 0.50 & 0.60 \\
   & CCS                & \cmark & 0.59 & 0.56 & 0.60 & 0.54 & 0.57 \\
   & \textbf{TDGNet (Ours)} & \cmark & \textbf{0.72} & \textbf{0.65} & \textbf{0.64} & \textbf{0.72} & \textbf{0.68} \\
\midrule
\multirow{8}{*}{\textbf{Dream-7B-Instruct}}
  & \multicolumn{7}{c}{\textbf{Output-Based}} \\
\cmidrule(l){2-8}
  & Semantic Entropy   & \xmark & 0.59 & 0.50 & 0.68 & 0.69 & 0.62 \\
  & Lexical Similarity & \xmark & \underline{0.71} & 0.68 & \underline{0.71} & 0.67 & \underline{0.69} \\
  & LN-Entropy         & \xmark & 0.57 & 0.59 & 0.52 & 0.53 & 0.55 \\
  & Perplexity         & \xmark & 0.52 & 0.57 & 0.51 & 0.54 & 0.54 \\
\cmidrule(l){2-8}
  & \multicolumn{7}{c}{\textbf{Latent-Based}} \\
\cmidrule(l){2-8}
  & EigenScore         & \xmark & 0.68 & 0.55 & 0.63 & \underline{0.70} & 0.64 \\
  & TSV                & \cmark & 0.68 & \underline{0.71} & 0.43 & 0.50 & 0.58 \\
  & CCS                & \cmark & 0.59 & 0.56 & 0.64 & 0.60 & 0.60 \\
   & \textbf{TDGNet (Ours)} & \cmark & \textbf{0.72} & \textbf{0.72} & \textbf{0.74} & \textbf{0.74} & \textbf{0.73} \\
\bottomrule
\end{tabular}
\end{table*}
\subsection{Problem Formulation}
\label{sec:method_problem}

We consider a Diffusion Language Model (D-LLM) that generates a response sequence $\mathbf{y}_0 = (y^1, \dots, y^n)$ conditioning on an input prompt $\mathbf{x}$. Unlike sequential auto-regressive generation, D-LLMs generate text via a global, iterative refinement process~\citep{li2022diffusion}. The generation commences from a fully masked sequence $\mathbf{y}_T$ (where all positions are initialized to a special mask token)~\citep{sahoo2024simple}. Over $T$ discrete timesteps, the model progressively denoises this sequence to produce a \textit{decoding trajectory} of intermediate states $\mathcal{T} = \{\mathbf{y}_T, \mathbf{y}_{T-1}, \dots, \mathbf{y}_0\}$~\citep{nie2025lldm}. Each transition is governed by a learned conditional distribution $p_\theta(\mathbf{y}_{t-1} | \mathbf{y}_t, \mathbf{x})$, where the model utilizes bidirectional attention to refine the semantic structure of the entire sequence simultaneously during the iterative decoding process~\citep{gong2024diffullama,zhang2025surveyparallel}.

In this context, we frame hallucination detection as evaluating the factual validity of the generated content. Let $h \in \{0,1\}$ be the ground-truth label where $h=1$ indicates a hallucination. The goal is to learn a binary classifier $f$ from a hypothesis space $\mathcal{H}$ that minimizes the classification risk for the response $\mathbf{y}_0$, subject to the constraint that $\mathbf{y}_0$ is the result of the model's reverse diffusion process:
\begin{equation}
\label{eq:optimization_objective}
\min_{f \in \mathcal{H}} \mathcal{L}(h, f(\mathbf{x}, \mathbf{y}_0)), \quad \text{s.t.} \quad \mathbf{y}_0 \sim \prod_{t=T}^{1} p_\theta(\mathbf{y}_{t-1} \mid \mathbf{y}_t, \mathbf{x})
\end{equation}
where $\mathcal{L}(\cdot)$ is the cross-entropy loss. This formulation explicitly links the detection task to the generative dynamics, acknowledging that while the target $\mathbf{y}_0$ is a static sequence, its validity is an artefact of the specific \textit{generative path} $p_\theta$.

However, applying conventional detection paradigms to this diffusion-based formulation is hindered by architectural mismatches. Regarding \textbf{output-based approaches}, the primary obstacle is \textit{likelihood intractability}: unlike AR models, D-LLMs lack a direct single-pass likelihood $p(\mathbf{y}|\mathbf{x})$, rendering perplexity-based metrics computationally expensive~\citep{sahoo2024simple}. Conversely, \textbf{latent-based strategies} struggle with \textit{temporal information loss}. Because they typically analyze only the final state $\mathbf{y}_0$ or a single snapshot, these methods discard rich validity signals, such as "semantic drift" or "persistent error", that manifest exclusively as dynamic instabilities within the intermediate trajectory terms $p_\theta(\mathbf{y}_{t-1} \mid \mathbf{y}_t)$~\citep{chang2025tracedet,wang2025time}.


\subsection{Capturing Bidirectional Dependencies and Evolution via Temporal Graph Modeling}
\label{sec:method_reformulation}

To effectively detect hallucinations in D-LLMs, a representation must capture two fundamental generative properties: the \textit{global context} facilitated by bidirectional attention and the \textit{iterative refinement} of semantic content over time. Standard static graph approaches fail to capture the latter, while sequence-only methods miss the former. Therefore, we propose to model the computational trace not as a fixed snapshot, but as a Temporal Dynamic Graph, explicitly encoding the generation as an evolving web of token dependencies.


\textbf{Dynamic Graph Construction.} 
Formally, we define the diffusion trace as a discrete-time sequence of attributed graphs $\mathcal{G} = \{G^{(0)}, \dots, G^{(T)}\}$. Each snapshot $G^{(t)} = (V, \mathcal{E}^{(t)}, \mathbf{X}_V^{(t)}, \mathbf{X}_E^{(t)})$ captures the structural state of the generation at denoising step $t$.
First, the {Vertex Set ($V$)} consists of the $N$ tokens in the sequence $\mathbf{y}$. Unlike AR generation, the node set remains fixed across the trajectory, serving as the spatial anchor for temporal evolution. 
Second, the {Temporal Edge Set ($\mathcal{E}^{(t)}$)} represents the dynamic connectivity. Leveraging the model's {bidirectional attention}, we define a directed \emph{message} edge $(j,i)$ at step $t$ if token $i$ significantly attends to token $j$:

\begin{equation}
\mathcal{E}^{(t)} = \left\{(j,i) \,\middle|\, \frac{1}{H} \sum_{h=1}^H \mathbf{A}_{ij}^{(t, L, h)} > \tau \right\}
\end{equation}

where $\tau$ is a sparsity threshold. This construction allows the graph topology to evolve naturally, capturing phenomena such as the emergence of semantic cliques or the dissolution of random noise connections.

To unify heterogeneous signals, we {enrich} this structure with time-varying features. {Node Attributes} $\mathbf{X}_V^{(t)}$ are derived from residual stream activations, projected to $\mathbb{R}^{d_v}$ to capture instantaneous semantic confidence and context. {Edge Attributes} $\mathbf{X}_E^{(t)}$ are derived from the {averaged attention scores} across heads, preserving interaction strength while maintaining computational efficiency.

By formulating the diffusion trace as a dynamic graph sequence $\mathcal{G}$, our approach unifies the structural analysis of token interactions with the temporal analysis of their evolution. This representation allows the model to distinguish between \textit{healthy refinement} (where dependencies stabilize) and \textit{hallucinatory collapse} (where dependencies drift or lock onto irrelevant tokens or noisy patterns).

This reformulation is necessitated by the path-dependent nature of D-LLM hallucinations. Unlike static errors, hallucinations often emerge through semantic drift during the denoising process ($t=T \to 0$), a phenomenon invisible to single-step analysis~\citep{chang2025tracedet}. By defining the input as a dynamic sequence $\mathcal{G}$, we capture this structural evolution, allowing the model to distinguish between stable grounding and transient instabilities, thus motivating the temporal architecture detailed in the following section.

\begin{table}[t]
\centering
\small
\caption{Comparison of graph-based variants for LLaDA. The highest score is \textbf{bolded} and the second highest is \underline{underlined}.}
\label{tab:graph_variations_combined}
\resizebox{\columnwidth}{!}{%
\begin{tabular}{l c r r r r r}
\toprule
\textbf{Variant} & \textbf{SS} & \textbf{Math} & \textbf{CSQA} & \textbf{HotpotQA} & \textbf{TriviaQA} & \textbf{Ave} \\
\midrule
CHARM — Decoding step 0     & \checkmark & 0.63 & 0.58 & 0.58 & 0.65 & 0.61 \\
CHARM — Decoding step T/4   & \checkmark & 0.65 & 0.60 & 0.61 & 0.68 & 0.63 \\
CHARM — Decoding step T/2   & \checkmark & \underline{0.70} & \underline{0.63} & \underline{0.63} & \underline{0.69} & \underline{0.66} \\
CHARM — Decoding step T     & \checkmark & 0.65 & 0.63 & 0.62 & 0.68 & 0.64 \\
\textbf{TDGNet (Ours)}      & \checkmark & \textbf{0.72} & \textbf{0.65} & \textbf{0.64} & \textbf{0.72} & \textbf{0.68} \\
\bottomrule
\end{tabular}%
}
\end{table}

\subsection{Hallucination Detection via Dynamic Trace Modeling}
\label{sec:method_details}

To classify the diffusion trajectory constructed in Section~\ref{sec:method_reformulation}, we introduce the Temporal Dynamic Graph Network (TDGNet). This architecture processes the sequence of attention graphs $\{\mathcal{G}^{(t)}\}_{t=T}^{0}$ via a three-stage pipeline, which consists of spatial aggregation, temporal memory update, and trajectory readout, to distinguish between factual and hallucinated generation paths.



First, to capture the local structural context at each diffusion step $t$, we employ a Message Passing Neural Network (MPNN)~\citep{gilmer2017mpnn}. This module aggregates information from the attention neighborhood $\mathcal{N}_i^{(t)}$. For each token $i$, we compute an aggregated message vector $\bar{\mathbf{m}}_i^{(t)}$ that summarizes the information flow from its neighbors:
\begin{equation}
\label{eq:message_passing}
\bar{\mathbf{m}}_i^{(t)} = \frac{1}{|\mathcal{N}^{(t)}(i)|} \sum_{j \in \mathcal{N}^{(t)}(i)} \psi \left( \mathbf{h}_j^{(t)}, \mathbf{h}_i^{(t)}, \mathbf{e}_{ji}^{(t)} \right)
\end{equation}
where $\mathbf{h}_i^{(t)}$ and $\mathbf{h}_j^{(t)}$ are the node features, $\mathbf{e}_{ji}^{(t)}$ are the edge features, and $\psi(\cdot)$ is a learnable message function implemented as a multi-layer perceptron (MLP). This spatial aggregation ensures that the model captures the instantaneous relational structure of the generation before processing its temporal evolution through recurrence.


Subsequently, to track how these structural roles evolve, we maintain a persistent memory state $\mathbf{s}_i^{(t)}$ for each token. We employ a Gated Recurrent Unit (GRU) to transition the memory state from step $t+1$ to $t$:
\begin{equation}
\label{eq:gru_update}
\mathbf{s}_i^{(t)} = \text{GRU}\left(\bar{\mathbf{m}}_i^{(t)}, \mathbf{s}_i^{(t+1)}\right)
\end{equation}
This recurrence allows the network to distinguish between consistent grounding and transient noise by integrating the sequence of spatial messages $\bar{\mathbf{m}}_i^{(t)}$ into a continuous "running belief" $\mathbf{s}_i^{(t)}$ representation over time.

Finally, to produce a robust prediction, we employ a Temporal Attention Mechanism that prioritizes informative denoising phases over noisy initialization. We compute a trajectory-aware embedding $\mathbf{z}_i$ for each token by attending over its own memory history:
\begin{equation}
\label{eq:temp_attn}
\begin{split}
\mathbf{z}_i &= \sum_{t=0}^{T} \alpha_i^{(t)} \cdot \text{Linear}(\mathbf{s}_i^{(t)}), \\
\text{where} \quad \alpha_i^{(t)} &= \frac{\exp(\mathbf{q}^\top \mathbf{s}_i^{(t)})}{\sum_{k=0}^T \exp(\mathbf{q}^\top \mathbf{s}_i^{(k)})}
\end{split}
\end{equation}
Here, $\text{Linear}(\cdot)$ denotes a linear projection layer, and $\mathbf{q}$ is a learnable query vector. This yields a final representation $\mathbf{z}_i$ that summarizes the token's structural validity across the entire trajectory of the generation process.

\textbf{Learning Objective.}
For the primary task of response-level hallucination detection, we apply global mean-pooling over the token embeddings to obtain a graph representation $\mathbf{g} = \frac{1}{N} \sum_{i} \mathbf{z}_i$. This is passed to a binary classifier $f_\phi$ to predict the probability $\hat{p}$. The model is trained end-to-end by minimizing the binary cross-entropy loss against the ground truth labels $h \in \{0, 1\}$ indicating response validity:
\begin{equation}
\mathcal{L} = - \frac{1}{|\mathcal{D}|} \sum_{(\mathbf{x}, \mathbf{y}) \in \mathcal{D}} \left[ h \log \hat{p} + (1-h) \log (1-\hat{p}) \right]
\end{equation}
where $\mathcal{D}$ represents the training dataset of prompt-response pairs and their corresponding ground-truth labels.



\subsection{Implementation Details}
\label{sec:method_implementation}

We instantiate the TDGNet framework as a two-stage pipeline comprising a \textbf{Temporal Graph Encoder} $g_\theta$ and a \textbf{Hallucination Predictor} $f_\phi$.

\textbf{(a) Temporal Graph Encoder Configuration.}
\begin{itemize}
    \item \textit{Graph Sparsity Threshold ($\tau$):} To robustly define the edge set $\mathcal{E}^{(t)}$, we employ a threshold $\tau$ rather than a Top-K constraint. This allows the graph topology to evolve naturally: distinct semantic cliques can emerge at later denoising stages without being artificially constrained, while the dense, uniform noise characteristic of early diffusion steps is effectively filtered out.

    \item \textit{Efficient Trajectory Sampling:} While TDGNet is formulated over the full denoising trace $\{G^{(t)}\}_{t=0}^{T}$ (processed in denoising order $t=T \rightarrow 0$), processing every step incurs high redundancy and computational cost. To balance fidelity with efficiency, we sample a strategic set of keyframes $\mathcal{T}_{\text{key}}=\{T,\lfloor T/2\rfloor,\lfloor T/4\rfloor,0\}$. This sequence captures the critical phases of generation: initial structural formation ($t=T$), semantic construction ($t=\lfloor T/2\rfloor$), and late-stage refinement ($t=0$). Crucially, this does not reduce the model to an ensemble of static snapshots; the recurrent memory states persist and update sequentially across these keyframes, allowing TDGNet to model the \textit{evolutionary dynamics} of the hallucination even with sparse temporal sampling.

    \item \textit{Feature Initialization:} Node features $\mathbf{h}_i^{(t)}$ are initialized from projected final-layer hidden states ($\mathbb{R}^{4096} \to \mathbb{R}^{128}$). The memory module uses a GRU with state dimension $d_m=128$, and the temporal attention query $\mathbf{q}$ is similarly set to $d_m=128$.
\end{itemize}

\textbf{(b) Dual-Task Readout Heads.} 
TDGNet supports both global and granular detection via specialized heads:
\begin{enumerate}
    \item \textbf{Response-Level (Graph Classification):} For sequence-level detection, we apply global mean-pooling over the final temporal node embeddings $\mathbf{z}_i$ to obtain $\mathbf{g}$, which is passed to a binary classifier $f_{\text{seq}}(\mathbf{g})$.
    
    \item \textbf{Token-Level (Node Classification):} For fine-grained localization, we bypass pooling and feed the learned individual node embeddings $\mathbf{z}_i$ into a shared token classifier $f_{\text{token}}(\mathbf{z}_i)$. This allows the model to pinpoint specific structural anomalies (e.g., local attention drift) without losing any granular resolution.
\end{enumerate}

\section{Experiments}
\begin{table}[t]
    \centering
    \caption{\textbf{Token-Level Hallucination Detection (AUROC).} Comparison of our dynamic method (TDGNet) and static graph baseline (CHARM) against standard baselines on Natural Questions (NQ) and TriviaQA. Best results are \textbf{bolded}, second best are \underline{underlined}.}
    \label{tab:token_results}
    \resizebox{\linewidth}{!}{
    \begin{tabular}{l c c}
        \toprule
        \textbf{Method} & \textbf{Natural Questions} & \textbf{TriviaQA} \\
        \midrule
        Graph Degree (Centrality) & 0.43 & 0.46 \\
        Source Attribution & 0.52 & 0.42 \\
        Token Probability & 0.60 & 0.68 \\
        Predictive Entropy & 0.69 & 0.71 \\
        \midrule
        CHARM (Static Graph) & \underline{0.84} & \underline{0.83} \\
        \textbf{TDGNet (Ours)} & \textbf{0.87} & \textbf{0.85} \\
        \bottomrule
    \end{tabular}
    }
\end{table}

\subsection{Setup}
\label{sec:exp_setup}

\textbf{Datasets.}
We evaluate hallucination detection on five QA-style benchmarks with varying reasoning demands:
\textbf{Math} (numerical reasoning, constructed following~\citep{chuang2024lookback}),
\textbf{CommonsenseQA (CSQA)} (commonsense multiple choice)~\citep{talmor2019commonsenseqa},
\textbf{HotpotQA} (multi-hop reasoning with supporting context)~\citep{yang2018hotpotqa},
\textbf{TriviaQA} (open-domain factoid QA)~\citep{joshi2017triviaqa}, and
\textbf{Natural Questions (NQ)} (factoid QA)~\citep{kwiatkowski2019natural}.
While all datasets are utilized for response-level detection, we specifically employ \textbf{NQ} and \textbf{TriviaQA} for the fine-grained \textit{token-level localization} experiments (Section~\ref{sec:main_results}), as their factoid nature supports precise span-level verification. Dataset construction, labeling, and splits follow Section~\ref{app:dataset_details}; all methods use identical train/test partitions.

\textbf{Baseline Methods.}
We compare TDGNet against a comprehensive suite of detectors across three paradigms, prioritizing methods that facilitate rigorous comparison on a unified benchmark (selection criteria in Appendix~\ref{app:baseline_criteria}). For \textbf{response-level detection}, we evaluate \textbf{output-based} methods operating on prediction statistics, including Semantic Entropy~\citep{kuhn2023semantic,kossen2024semantic}, Length-Normalized (LN) Entropy and Perplexity~\citep{lin2023generating}, and Lexical Similarity~\citep{manakul2023selfcheckgpt}; \textbf{latent-based} methods probing internal representations, such as EigenScore~\citep{chen2024inside}, TSV~\citep{park2025tsv}, and consistency probes like CCS~\citep{azaria2023internal,orgad2024llmsknow}; and \textbf{graph-based} methods represented by CHARM~\citep{frasca2025charm}, evaluated via an in-house implementation. For \textbf{token-level localization}, since standard response metrics suffer from a \textit{granularity mismatch} (broadcasting a single scalar dilutes local error signals), we compare against \textbf{native token-wise} baselines that test distinct validity hypotheses: (1) \textbf{Predictive Entropy}~\citep{kuhn2023semantic} (testing if hallucinations correlate with local uncertainty); (2) \textbf{Graph Degree Centrality} (testing the "Hub Hypothesis" that grounded tokens act as attention hubs); and (3) \textbf{Source Attribution}~\citep{bahdanau2014neural,vaswani2017attention} (testing the "Context Hypothesis" that factual tokens maintain strong attention to the prompt).

\begin{figure}[t]
    \centering
    \includegraphics[width=.9\linewidth]{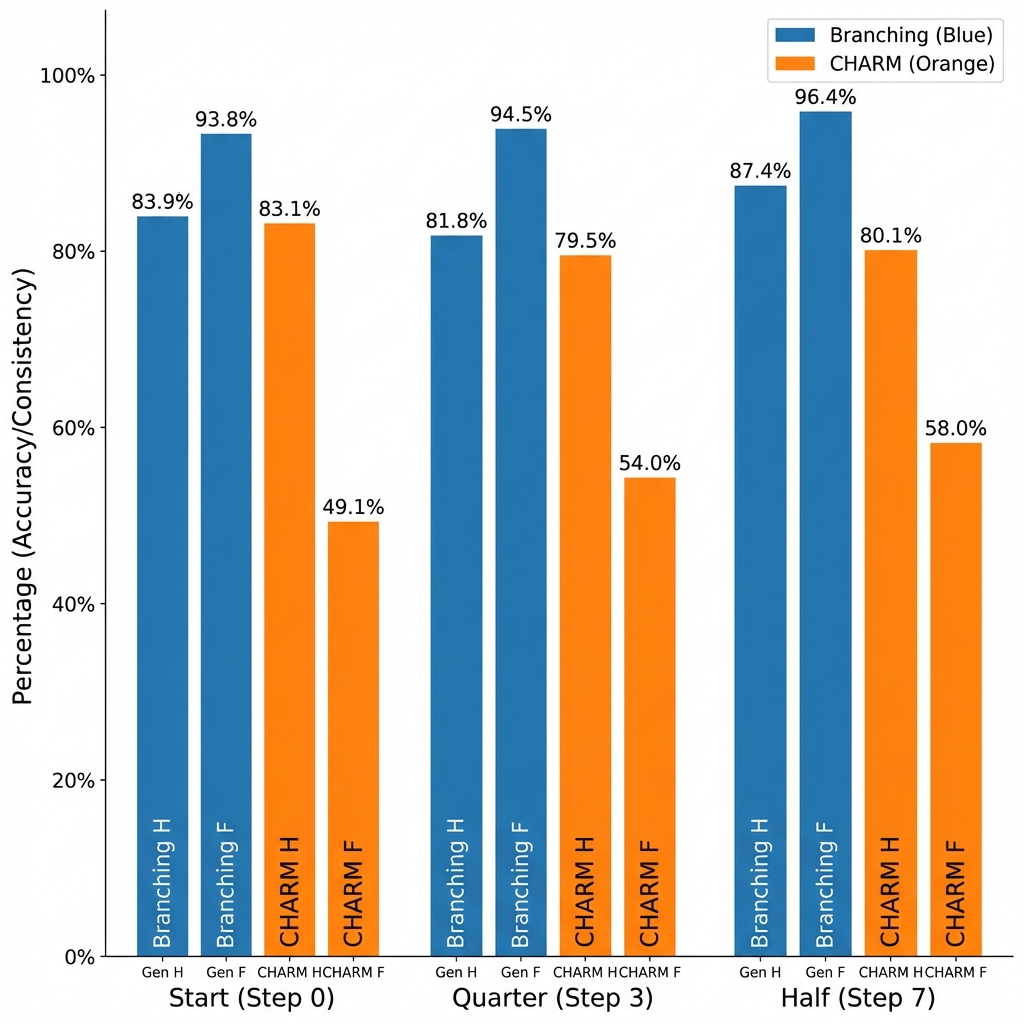}
\caption{\textbf{Temporal Consistency Analysis.} We compare the inherent validity signal in the data (Blue, "Branching Fate") against the accuracy of static snapshots (Orange, CHARM). \textbf{Key Insight:} While the data contains strong early signals ($>81\%$ consistency), static models fail to capture them (dropping to 49.1\% accuracy). This gap mathematically motivates TDGNet: the signal exists in the trajectory, but requires temporal modeling to extract.}
\vspace{-15pt}
    \label{fig:fate_barchart}
\end{figure}

\textbf{Models.}  We adopt \textbf{LLaDA-8B-Instruct} and \textbf{Dream-7B-Instruct} as representative open-source diffusion LMs~\citep{nie2025lldm,ye2025dream7b}. Using both models enables a unified comparison across different architectures and noise schedules. For graph methods, we construct attention graphs from the final transformer layer using a standardized decoding configuration (\texttt{gen\_length}=32, \texttt{steps}=16); we refer readers to Appendix~\ref{app:ablation} for a detailed sensitivity analysis regarding these hyperparameters.

Notably, given the limited landscape of open-source D-LLMs, we include Dream-7B despite its known tendency toward \textit{structural degradation} (mode collapse) in zero-shot settings. To ensure this does not confound our evaluation, we employ a temperature-scaled curation strategy that separates \textit{generation coherence} (fluency) from \textit{factuality} (hallucination). This protocol, detailed in Appendix~\ref{subsec:dream_instability}, ensures that our benchmark measures the detector's ability to identify plausible falsehoods rather than syntactic failures.

\textbf{Graph-based variants.}
To disentangle the contribution of temporal modeling from graph-based structural analysis, we introduce a set of \textbf{static graph variants}. This comparison aims to verify whether detection performance stems merely from the graph representation itself or specifically from tracking its \textit{evolution} over the denoising process. We evaluate static models, based on ~\citep{frasca2025charm}, operating on single snapshots extracted at key intervals $t \in \{0, T/4, T/2, T\}$, contrasting them with our dynamic TDGNet, which consumes the trajectory sequence.

\textbf{Evaluation protocol.}
Following prior work~\citep{kuhn2023semantic,manakul2023selfcheckgpt,park2025tsv,chang2025tracedet}, we report AUROC as the primary metric, selecting models on validation data. Table~\ref{tab:baseline_vs_graph_combined} summarizes performance across all benchmarks. TDGNet achieves the best average AUROC on LLaDA-8B and strong results on Dream-7B, confirming the benefit of temporal modeling over output-, latent-, and static baselines.

\begingroup
\setlength{\abovecaptionskip}{2pt}
\setlength{\belowcaptionskip}{0pt} 
\begin{table}[t]
\centering
\caption{\textbf{Graph Utility Ablation.} The drastic drop without edges ($\mathcal{N}(i)=\emptyset$) confirms structural aggregation is essential.}
\label{tab:ablation_graph_utility}

\setlength{\tabcolsep}{0pt}
\begin{tabular*}{\columnwidth}{@{\extracolsep{\fill}} l c c c}
\toprule
 & \multicolumn{2}{c}{\textbf{Performance (AUROC)}} & \\
\cmidrule(lr){2-3}
\textbf{Dataset} & \textbf{TDGNet} & \textbf{No Graph} & \textbf{$\Delta$ Drop} \\
\midrule
CSQA      & 0.65 & 0.40 & -0.25 \\
TriviaQA  & 0.72 & 0.43 & -0.29 \\
HotpotQA  & 0.64 & 0.48 & -0.16 \\
Math      & 0.72 & 0.49 & -0.23 \\
\bottomrule
\end{tabular*}
\end{table}
\endgroup

\subsection{Main Results}
\label{sec:main_results}

\textbf{Response-Level Detection.}
Table~\ref{tab:baseline_vs_graph_combined} presents a comprehensive comparison of TDGNet against baseline hallucination detection methods across two D-LLMs and four QA benchmarks. TDGNet achieves the highest performance in all settings, outperforming the second-strongest baseline (TSV) by 8.5\% AUROC on LLaDA-8B-Instruct and Lexical Similarity by 5.3\% on Dream-7B-Instruct. These consistent gains exhibit the value of exploiting the multi-step evolution of the denoising trajectory rather than relying solely on static output statistics or isolated latent snapshots.

Among the baselines, Output-based approaches suffer from poor consistency. Semantic Entropy achieves competitive performance on LLaDA but collapses to 50.3\% on Dream-7B-Instruct (CSQA), underscoring poor robustness when hallucinations do not manifest as high entropic uncertainty. Furthermore, these methods require multiple generations, imposing computational overhead unsuitable for production. Latent-based methods capture hidden-state geometry and show more competitive results, but they remain sensitive to task distribution. CCS achieves 64.3\% AUROC on HotpotQA yet fluctuates significantly, dropping to near-random performance on TriviaQA. This inconsistency highlights the limitation of seeking a universal "truth direction" in static embeddings without accounting for temporal dynamics.

To isolate the contribution of temporal modeling, Table~\ref{tab:graph_variations_combined} compares our method against static CHARM variants. While static graphs at mid-trajectory ($t=T/2$) perform better than early or late snapshots, they consistently underperform TDGNet. This demonstrates that hallucination signals are distributed across the trajectory, and that effective detection requires the trajectory-level aggregation that our temporal framework provides.

\textbf{Token-Level Localization.}
Beyond sequence-level detection, Table~\ref{tab:token_results} validates the granular capabilities of our architecture. Standard uncertainty metrics like Predictive Entropy suffer from a granularity mismatch, achieving only $\sim$70\% AUROC, as broadcasting a single sequence-level score dilutes local error signals. In contrast, TDGNet leverages native node embeddings to outperform the static CHARM baseline by 2.1\% on the TriviaQA benchmark. This confirms that the temporal memory mechanism does not merely aggregate signals but refines the structural boundaries of hallucinations, allowing for the precise localization of errors that static topology alone fails to isolate.

\begin{table}[t]
    \centering
    \caption{Cross-dataset zero-shot AUROC on LLaDA: Grouped by Method. Best results are \textbf{bolded}, second best are \underline{underlined}.}
    \label{tab:transfer_method_grouped}
    \resizebox{\linewidth}{!}{%
        \begin{tabular}{llcccc}
            \toprule
            & & \multicolumn{4}{c}{\textbf{Target Dataset}} \\
            \cmidrule(lr){3-6}
            \textbf{Method} & \textbf{Source} & Math & CSQA & Hotpot & Trivia \\
            \midrule
            \multirow{4}{*}{CCS} 
            & Math   & 0.59 & 0.52 & 0.53 & 0.50 \\
            & CSQA   & 0.51 & 0.56 & 0.50 & 0.54 \\
            & Hotpot & 0.54 & 0.55 & 0.60 & 0.59 \\
            & Trivia & 0.52 & 0.51 & 0.51 & 0.54 \\
            \midrule
            \multirow{4}{*}{\textbf{TDGNet (Ours)}} 
            & Math   & \textbf{0.72} & 0.49 & \underline{0.62} & 0.55 \\
            & CSQA   & 0.55 & \textbf{0.65} & 0.59 & 0.58 \\
            & Hotpot & \underline{0.61} & \underline{0.59} & \textbf{0.64} & \underline{0.62} \\
            & Trivia & 0.57 & 0.55 & 0.57 & \textbf{0.72} \\
            \bottomrule
        \end{tabular}%
    }
\end{table}

\begin{table}[t]
    \centering
\caption{\textbf{Efficiency Comparison.} TDGNet incurs marginal overhead over single-pass baselines, whereas sampling-based methods (Semantic Entropy, Lexical Similarity, EigenScore) are significantly slower due to multiple generations.}
    \label{tab:efficiency}
    \resizebox{\linewidth}{!}{%
        \begin{tabular}{lccc}
            \toprule
            \textbf{Method} & \textbf{Generations / Query} & \textbf{Overhead / Query} & \textbf{Time (100 Queries)} \\
            \midrule
            Perplexity          & 1 & $\sim$66 ms & 426 s \\
            LN-Entropy          & 1 & $\sim$66 ms & 426 s \\
            CCS                 & 1 & $\sim$96 ms & 430 s \\
            TSV                 & 1 & $\sim$66 ms & 426 s \\
            \midrule
            Semantic Entropy    & 5 & $\sim$10 ms & 2101 s \\
            Lexical Similarity  & 5 & $\sim$1 ms & 2100 s \\
            EigenScore          & 5 & $\sim$14 ms & 2102 s \\
            \midrule
            \textbf{TDGNet (Ours)} & 1 & $\sim$160 ms & \textbf{442 s} \\
            \bottomrule
        \end{tabular}%
    }
\end{table}

\subsection{Analysis}

\textbf{Importance of Message-Passing Structure.} 
Hallucination detection requires capturing the relational structure of attention, which latent-based methods often discard. By utilizing the spatial aggregation mechanism defined in Eq.~(\ref{eq:message_passing}), our architecture updates each token's embedding based on its attention-induced neighborhood $\mathcal{N}(i)$. This allows the model to capture pathological patterns like inconsistent grounding and unstable recirculation. Ablating graph edges ($\mathcal{N}(i) = \emptyset$) invalidates this aggregation, reducing the model to a "Bag of Token Confidences" where prediction relies solely on isolated hidden states. This causes performance to collapse to near-random chance in all datasets (Table~\ref{tab:ablation_graph_utility}), validating that isolated node features are insufficient; effective detection requires structural aggregation to unify isolated heuristics into a learned, end-to-end framework.

\textbf{Temporal Dynamics.} \label{sec:fate_analysis}
While CHARM demonstrates the value of attention graphs, its restriction to single-step snapshots limits its ability to model the diffusion process. Static variants peak at mid-trajectory (0.6627 AUROC) yet consistently trail our dynamic approach (0.6814) (Table~\ref{tab:graph_variations_combined}). Figure~\ref{fig:fate_barchart} illustrates the mechanism behind this gap: while the trajectory itself holds consistent validity signals ($>83\%$ consistency, Blue bars), static models fail to capture them, dropping to 49.1\% accuracy at Step 0. This disparity between the \textit{inherent} data signal and the \textit{extracted} signal confirms that validity cues are not localized to any single snapshot, but are instead \textit{distributed} throughout the denoising history. TDGNet bridges this gap by trajectory-level aggregation, allowing it to distinguish true structural stability from the transient artifacts that confound static methods.

\textbf{Generalization and Cross-dataset Analysis.} Cross-dataset zero-shot experiments (Table~\ref{tab:transfer_method_grouped}) demonstrate that graph-based detection generalizes more robustly than latent-space methods. TDGNet maintains meaningful transfer (0.55--0.62 AUROC) across diverse reasoning types, while CCS shows near-random performance on distant pairs (0.50--0.54 AUROC). Within knowledge-retrieval tasks, transfer is strong (0.57--0.62), reflecting shared patterns for entity grounding and evidence verification. The strong performance of Hotpot as a source dataset (0.59--0.62 average transfer) suggests that multi-hop reasoning patterns generalise, indicating that core hallucination signatures, such as attention inconsistency and semantic drift, are partially invariant to task-specific reasoning modes (see Appendix~\ref{subsec:ablation_generalization}).

\textbf{Efficiency.} Inference efficiency is critical for D-LLMs. While consistency-based methods (e.g., Semantic Entropy) inflate latency by requiring multiple generations ($S=5$), TDGNet leverages a \textit{single} denoising trajectory. Table~\ref{tab:efficiency} confirms a $\sim$4.8$\times$ speedup (442s) over Semantic Entropy (2101s). The marginal overhead compared to Perplexity (442s vs 426s) is negligible relative to the cost of additional sampling, allowing TDGNet to combine single-pass speed with the robustness of ensemble approaches.
\label{sec:res}

\section{Conclusion}

TDGNet reframes hallucination detection for diffusion language models as learning over the \emph{evolving} attention-graph trace, operationalizing the principle that factuality is a \emph{trajectory property} rather than solely an end-state outcome. By utilizing token-level message passing and persistent memories to accumulate weak validity cues across denoising \emph{stages}, our framework consistently improves AUROC over output-based, latent-based, and static graph baselines on both LLaDA-8B and Dream-7B, while simultaneously enabling fine-grained localization of hallucinated spans. Analyses confirm that these gains stem from explicitly modeling diffusion-specific dynamics, such as self-correction, semantic drift, and persistent errors, allowing the detector to capture early warning signs and late-stage repair effects that static snapshots miss. Ultimately, by operating in a single pass with modest overhead, TDGNet offers a practical path toward scalable hallucination monitoring and a foundation for future mitigation in D-LLMs.

\section*{Impact Statement}

This paper presents work whose goal is to advance the field of machine learning by improving the reliability and trustworthiness of Diffusion Language Models. By introducing a mechanism to detect and localize hallucinations, our research aims to mitigate the risks associated with the generation of misinformation and supports the safer deployment of generative AI systems in real-world applications. We believe this work contributes positively to the responsible development of Large Language Models, and there are no specific negative societal consequences that we feel must be highlighted here.

\bibliography{example_paper}

\begin{thebibliography}{52}
\providecommand{\natexlab}[1]{#1}
\providecommand{\url}[1]{\texttt{#1}}
\expandafter\ifx\csname urlstyle\endcsname\relax
  \providecommand{\doi}[1]{doi: #1}\else
  \providecommand{\doi}{doi: \begingroup \urlstyle{rm}\Url}\fi

\bibitem[Achiam et~al.(2023)Achiam, Adler, Agarwal, Ahmad, Akkaya, Aleman, Almeida, Altenschmidt, Altman, Anadkat, et~al.]{openai2023gpt4}
Achiam, J., Adler, S., Agarwal, S., Ahmad, L., Akkaya, I., Aleman, F.~L., Almeida, D., Altenschmidt, J., Altman, S., Anadkat, S., et~al.
\newblock Gpt-4 technical report.
\newblock \emph{arXiv preprint arXiv:2303.08774}, 2023.

\bibitem[Azaria \& Mitchell(2023)Azaria and Mitchell]{azaria2023internal}
Azaria, A. and Mitchell, T.
\newblock The internal state of an llm knows when it's lying.
\newblock \emph{arXiv preprint arXiv:2304.13734}, 2023.

\bibitem[Bahdanau et~al.(2016)Bahdanau, Cho, and Bengio]{bahdanau2014neural}
Bahdanau, D., Cho, K., and Bengio, Y.
\newblock Neural machine translation by jointly learning to align and translate, 2016.
\newblock URL \url{https://arxiv.org/abs/1409.0473}.

\bibitem[Battaglia et~al.(2018)Battaglia, Hamrick, Bapst, Sanchez-Gonzalez, Zambaldi, Malinowski, Tacchetti, Raposo, Santoro, Faulkner, et~al.]{battaglia2018relational}
Battaglia, P.~W., Hamrick, J.~B., Bapst, V., Sanchez-Gonzalez, A., Zambaldi, V., Malinowski, M., Tacchetti, A., Raposo, D., Santoro, A., Faulkner, R., et~al.
\newblock Relational inductive biases, deep learning, and graph networks.
\newblock \emph{arXiv preprint arXiv:1806.01261}, 2018.

\bibitem[Brown et~al.(2020)Brown, Mann, Ryder, Subbiah, Kaplan, Dhariwal, Neelakantan, Shyam, Sastry, Askell, et~al.]{brown2020language}
Brown, T., Mann, B., Ryder, N., Subbiah, M., Kaplan, J.~D., Dhariwal, P., Neelakantan, A., Shyam, P., Sastry, G., Askell, A., et~al.
\newblock Language models are few-shot learners.
\newblock \emph{Advances in neural information processing systems}, 33:\penalty0 1877--1901, 2020.

\bibitem[Burns et~al.(2022)Burns, Ye, Klein, and Steinhardt]{burns2022ccs}
Burns, C., Ye, H., Klein, D., and Steinhardt, J.
\newblock Discovering latent knowledge in language models without supervision.
\newblock \emph{arXiv preprint arXiv:2212.03827}, 2022.

\bibitem[Chang et~al.(2025)Chang, Yu, Wang, Chen, Yu, Torr, and Gu]{chang2025tracedet}
Chang, S., Yu, J., Wang, W., Chen, Y., Yu, J., Torr, P., and Gu, J.
\newblock Tracedet: Hallucination detection from the decoding trace of diffusion large language models.
\newblock \emph{arXiv preprint arXiv:2510.01274}, 2025.

\bibitem[Chen et~al.(2024)Chen, Liu, Chen, Gu, Wu, Tao, Fu, and Ye]{chen2024inside}
Chen, C., Liu, K., Chen, Z., Gu, Y., Wu, Y., Tao, M., Fu, Z., and Ye, J.
\newblock Inside: Llms' internal states retain the power of hallucination detection.
\newblock \emph{arXiv preprint arXiv:2402.03744}, 2024.

\bibitem[Chern et~al.(2023)Chern, Chern, Chen, Yuan, Feng, Zhou, He, Neubig, Liu, et~al.]{chern2023factool}
Chern, I., Chern, S., Chen, S., Yuan, W., Feng, K., Zhou, C., He, J., Neubig, G., Liu, P., et~al.
\newblock Factool: Factuality detection in generative ai--a tool augmented framework for multi-task and multi-domain scenarios.
\newblock \emph{arXiv preprint arXiv:2307.13528}, 2023.

\bibitem[Chuang et~al.(2024)Chuang, Qiu, Hsieh, Krishna, Kim, and Glass]{chuang2024lookback}
Chuang, Y.-S., Qiu, L., Hsieh, C.-Y., Krishna, R., Kim, Y., and Glass, J.
\newblock Lookback lens: Detecting and mitigating contextual hallucinations in large language models using only attention maps.
\newblock \emph{arXiv preprint arXiv:2407.07071}, 2024.

\bibitem[Du et~al.(2024)Du, Xiao, and Li]{du2024haloscope}
Du, X., Xiao, C., and Li, S.
\newblock Haloscope: Harnessing unlabeled llm generations for hallucination detection.
\newblock \emph{Advances in Neural Information Processing Systems}, 37:\penalty0 102948--102972, 2024.

\bibitem[Frasca et~al.(2025)Frasca, Bar-Shalom, Ziser, and Maron]{frasca2025charm}
Frasca, F., Bar-Shalom, G., Ziser, Y., and Maron, H.
\newblock Neural message-passing on attention graphs for hallucination detection.
\newblock \emph{arXiv preprint arXiv:2509.24770}, 2025.

\bibitem[Ghazvininejad et~al.(2019)Ghazvininejad, Levy, Liu, and Zettlemoyer]{ghazvininejad2019mask}
Ghazvininejad, M., Levy, O., Liu, Y., and Zettlemoyer, L.
\newblock Mask-predict: Parallel decoding of conditional masked language models.
\newblock \emph{arXiv preprint arXiv:1904.09324}, 2019.

\bibitem[Gilmer et~al.(2017)Gilmer, Schoenholz, Riley, Vinyals, and Dahl]{gilmer2017mpnn}
Gilmer, J., Schoenholz, S.~S., Riley, P.~F., Vinyals, O., and Dahl, G.~E.
\newblock Neural message passing for quantum chemistry.
\newblock In \emph{International conference on machine learning}, pp.\  1263--1272. Pmlr, 2017.

\bibitem[Gong et~al.(2024)Gong, Agarwal, Zhang, Ye, Zheng, Li, An, Zhao, Bi, Han, et~al.]{gong2024diffullama}
Gong, S., Agarwal, S., Zhang, Y., Ye, J., Zheng, L., Li, M., An, C., Zhao, P., Bi, W., Han, J., et~al.
\newblock Scaling diffusion language models via adaptation from autoregressive models.
\newblock \emph{arXiv preprint arXiv:2410.17891}, 2024.

\bibitem[Hamilton et~al.(2017)Hamilton, Ying, and Leskovec]{hamilton2017graphsage}
Hamilton, W., Ying, Z., and Leskovec, J.
\newblock Inductive representation learning on large graphs.
\newblock \emph{Advances in neural information processing systems}, 30, 2017.

\bibitem[Joshi et~al.(2017)Joshi, Choi, Weld, and Zettlemoyer]{joshi2017triviaqa}
Joshi, M., Choi, E., Weld, D.~S., and Zettlemoyer, L.
\newblock Triviaqa: A large scale distantly supervised challenge dataset for reading comprehension.
\newblock \emph{arXiv preprint arXiv:1705.03551}, 2017.

\bibitem[Kipf(2016)]{kipf2017gcn}
Kipf, T.
\newblock Semi-supervised classification with graph convolutional networks.
\newblock \emph{arXiv preprint arXiv:1609.02907}, 2016.

\bibitem[Kossen et~al.(2024)Kossen, Han, Razzak, Schut, Malik, and Gal]{kossen2024semantic}
Kossen, J., Han, J., Razzak, M., Schut, L., Malik, S., and Gal, Y.
\newblock Semantic entropy probes: Robust and cheap hallucination detection in llms.
\newblock \emph{arXiv preprint arXiv:2406.15927}, 2024.

\bibitem[Kuhn et~al.(2023)Kuhn, Gal, and Farquhar]{kuhn2023semantic}
Kuhn, L., Gal, Y., and Farquhar, S.
\newblock Semantic uncertainty: Linguistic invariances for uncertainty estimation in natural language generation.
\newblock \emph{arXiv preprint arXiv:2302.09664}, 2023.

\bibitem[Kumar et~al.(2019)Kumar, Zhang, and Leskovec]{kumar2019jodie}
Kumar, S., Zhang, X., and Leskovec, J.
\newblock Predicting dynamic embedding trajectory in temporal interaction networks.
\newblock In \emph{Proceedings of the 25th ACM SIGKDD international conference on knowledge discovery \& data mining}, pp.\  1269--1278, 2019.

\bibitem[Kwiatkowski et~al.(2019)Kwiatkowski, Palomaki, Redfield, Collins, Parikh, Alberti, Epstein, Polosukhin, Devlin, Lee, et~al.]{kwiatkowski2019natural}
Kwiatkowski, T., Palomaki, J., Redfield, O., Collins, M., Parikh, A., Alberti, C., Epstein, D., Polosukhin, I., Devlin, J., Lee, K., et~al.
\newblock Natural questions: a benchmark for question answering research.
\newblock \emph{Transactions of the Association for Computational Linguistics}, 7:\penalty0 453--466, 2019.

\bibitem[Li et~al.(2022)Li, Thickstun, Gulrajani, Liang, and Hashimoto]{li2022diffusion}
Li, X., Thickstun, J., Gulrajani, I., Liang, P.~S., and Hashimoto, T.~B.
\newblock Diffusion-lm improves controllable text generation.
\newblock \emph{Advances in neural information processing systems}, 35:\penalty0 4328--4343, 2022.

\bibitem[Lin et~al.(2023)Lin, Trivedi, and Sun]{lin2023generating}
Lin, Z., Trivedi, S., and Sun, J.
\newblock Generating with confidence: Uncertainty quantification for black-box large language models.
\newblock \emph{arXiv preprint arXiv:2305.19187}, 2023.

\bibitem[Liu et~al.(2025)Liu, Chen, Ding, Song, Wang, Wu, and Wang]{liu2025agser}
Liu, Q., Chen, X., Ding, Y., Song, B., Wang, W., Wu, S., and Wang, L.
\newblock Attention-guided self-reflection for zero-shot hallucination detection in large language models.
\newblock In \emph{Proceedings of the 2025 Conference on Empirical Methods in Natural Language Processing}, pp.\  21016--21032, 2025.

\bibitem[Lou et~al.(2023)Lou, Meng, and Ermon]{lou2024discrete}
Lou, A., Meng, C., and Ermon, S.
\newblock Discrete diffusion modeling by estimating the ratios of the data distribution.
\newblock \emph{arXiv preprint arXiv:2310.16834}, 2023.

\bibitem[Manakul et~al.(2023)Manakul, Liusie, and Gales]{manakul2023selfcheckgpt}
Manakul, P., Liusie, A., and Gales, M.
\newblock Selfcheckgpt: Zero-resource black-box hallucination detection for generative large language models.
\newblock In \emph{Proceedings of the 2023 conference on empirical methods in natural language processing}, pp.\  9004--9017, 2023.

\bibitem[Min et~al.(2023)Min, Krishna, Lyu, Lewis, Yih, Koh, Iyyer, Zettlemoyer, and Hajishirzi]{min2023factscore}
Min, S., Krishna, K., Lyu, X., Lewis, M., Yih, W.-t., Koh, P., Iyyer, M., Zettlemoyer, L., and Hajishirzi, H.
\newblock Factscore: Fine-grained atomic evaluation of factual precision in long form text generation.
\newblock In \emph{Proceedings of the 2023 Conference on Empirical Methods in Natural Language Processing}, pp.\  12076--12100, 2023.

\bibitem[Mireshghallah et~al.(2022)Mireshghallah, Goyal, and Berg-Kirkpatrick]{mireshghallah2022mix}
Mireshghallah, F., Goyal, K., and Berg-Kirkpatrick, T.
\newblock Mix and match: Learning-free controllable text generation using energy language models.
\newblock \emph{arXiv preprint arXiv:2203.13299}, 2022.

\bibitem[Nie et~al.(2024)Nie, Hou, Lin, Zou, Yao, and Zhang]{nie2024facttest}
Nie, F., Hou, X., Lin, S., Zou, J., Yao, H., and Zhang, L.
\newblock Facttest: Factuality testing in large language models with finite-sample and distribution-free guarantees.
\newblock \emph{arXiv preprint arXiv:2411.02603}, 2024.

\bibitem[Nie et~al.(2025)Nie, Zhu, You, Zhang, Ou, Hu, Zhou, Lin, Wen, and Li]{nie2025lldm}
Nie, S., Zhu, F., You, Z., Zhang, X., Ou, J., Hu, J., Zhou, J., Lin, Y., Wen, J.-R., and Li, C.
\newblock Large language diffusion models.
\newblock \emph{arXiv preprint arXiv:2502.09992}, 2025.

\bibitem[Orgad et~al.(2024)Orgad, Toker, Gekhman, Reichart, Szpektor, Kotek, and Belinkov]{orgad2024llmsknow}
Orgad, H., Toker, M., Gekhman, Z., Reichart, R., Szpektor, I., Kotek, H., and Belinkov, Y.
\newblock Llms know more than they show: On the intrinsic representation of llm hallucinations.
\newblock \emph{arXiv preprint arXiv:2410.02707}, 2024.

\bibitem[Ouyang et~al.(2022)Ouyang, Wu, Jiang, Almeida, Wainwright, Mishkin, Zhang, Agarwal, Slama, Ray, et~al.]{ouyang2022training}
Ouyang, L., Wu, J., Jiang, X., Almeida, D., Wainwright, C., Mishkin, P., Zhang, C., Agarwal, S., Slama, K., Ray, A., et~al.
\newblock Training language models to follow instructions with human feedback.
\newblock \emph{Advances in neural information processing systems}, 35:\penalty0 27730--27744, 2022.

\bibitem[Pareja et~al.(2020)Pareja, Domeniconi, Chen, Ma, Suzumura, Kanezashi, Kaler, Schardl, and Leiserson]{pareja2020evolvegcn}
Pareja, A., Domeniconi, G., Chen, J., Ma, T., Suzumura, T., Kanezashi, H., Kaler, T., Schardl, T., and Leiserson, C.
\newblock Evolvegcn: Evolving graph convolutional networks for dynamic graphs.
\newblock In \emph{Proceedings of the AAAI conference on artificial intelligence}, volume~34, pp.\  5363--5370, 2020.

\bibitem[Park et~al.(2025)Park, Du, Yeh, Wang, and Li]{park2025tsv}
Park, S., Du, X., Yeh, M.-H., Wang, H., and Li, Y.
\newblock Steer llm latents for hallucination detection.
\newblock \emph{arXiv preprint arXiv:2503.01917}, 2025.

\bibitem[Rossi et~al.(2020)Rossi, Chamberlain, Frasca, Eynard, Monti, and Bronstein]{rossi2020tgn}
Rossi, E., Chamberlain, B., Frasca, F., Eynard, D., Monti, F., and Bronstein, M.
\newblock Temporal graph networks for deep learning on dynamic graphs.
\newblock \emph{arXiv preprint arXiv:2006.10637}, 2020.

\bibitem[Sahoo et~al.(2024)Sahoo, Arriola, Schiff, Gokaslan, Marroquin, Chiu, Rush, and Kuleshov]{sahoo2024simple}
Sahoo, S., Arriola, M., Schiff, Y., Gokaslan, A., Marroquin, E., Chiu, J., Rush, A., and Kuleshov, V.
\newblock Simple and effective masked diffusion language models.
\newblock \emph{Advances in Neural Information Processing Systems}, 37:\penalty0 130136--130184, 2024.

\bibitem[Sankar et~al.(2020)Sankar, Wu, Gou, Zhang, and Yang]{sankar2020dysat}
Sankar, A., Wu, Y., Gou, L., Zhang, W., and Yang, H.
\newblock Dysat: Deep neural representation learning on dynamic graphs via self-attention networks.
\newblock In \emph{Proceedings of the 13th international conference on web search and data mining}, pp.\  519--527, 2020.

\bibitem[Su et~al.(2024{\natexlab{a}})Su, Tang, Ai, Wang, Wu, and Liu]{sun2024redeep}
Su, W., Tang, Y., Ai, Q., Wang, C., Wu, Z., and Liu, Y.
\newblock Mitigating entity-level hallucination in large language models.
\newblock In \emph{Proceedings of the 2024 Annual International ACM SIGIR Conference on Research and Development in Information Retrieval in the Asia Pacific Region}, pp.\  23--31, 2024{\natexlab{a}}.

\bibitem[Su et~al.(2024{\natexlab{b}})Su, Wang, Ai, Hu, Wu, Zhou, and Liu]{su2024realtime}
Su, W., Wang, C., Ai, Q., Hu, Y., Wu, Z., Zhou, Y., and Liu, Y.
\newblock Unsupervised real-time hallucination detection based on the internal states of large language models.
\newblock \emph{arXiv preprint arXiv:2403.06448}, 2024{\natexlab{b}}.

\bibitem[Talmor et~al.(2019)Talmor, Herzig, Lourie, and Berant]{talmor2019commonsenseqa}
Talmor, A., Herzig, J., Lourie, N., and Berant, J.
\newblock Commonsenseqa: A question answering challenge targeting commonsense knowledge.
\newblock In \emph{Proceedings of the 2019 Conference of the North American Chapter of the Association for Computational Linguistics: Human Language Technologies, Volume 1 (Long and Short Papers)}, pp.\  4149--4158, 2019.

\bibitem[Touvron et~al.(2023)Touvron, Lavril, Izacard, Martinet, Lachaux, Lacroix, Rozi{\`e}re, Goyal, Hambro, Azhar, et~al.]{touvron2023llama}
Touvron, H., Lavril, T., Izacard, G., Martinet, X., Lachaux, M.-A., Lacroix, T., Rozi{\`e}re, B., Goyal, N., Hambro, E., Azhar, F., et~al.
\newblock Llama: Open and efficient foundation language models.
\newblock \emph{arXiv preprint arXiv:2302.13971}, 2023.

\bibitem[Trivedi et~al.(2019)Trivedi, Farajtabar, Biswal, and Zha]{trivedi2019dyrep}
Trivedi, R., Farajtabar, M., Biswal, P., and Zha, H.
\newblock Dyrep: Learning representations over dynamic graphs.
\newblock In \emph{International conference on learning representations}, 2019.

\bibitem[Vaswani et~al.(2017)Vaswani, Shazeer, Parmar, Uszkoreit, Jones, Gomez, Kaiser, and Polosukhin]{vaswani2017attention}
Vaswani, A., Shazeer, N., Parmar, N., Uszkoreit, J., Jones, L., Gomez, A.~N., Kaiser, {\L}., and Polosukhin, I.
\newblock Attention is all you need.
\newblock \emph{Advances in neural information processing systems}, 30, 2017.

\bibitem[Veli{\v{c}}kovi{\'c} et~al.(2017)Veli{\v{c}}kovi{\'c}, Cucurull, Casanova, Romero, Lio, and Bengio]{velickovic2018gat}
Veli{\v{c}}kovi{\'c}, P., Cucurull, G., Casanova, A., Romero, A., Lio, P., and Bengio, Y.
\newblock Graph attention networks.
\newblock \emph{arXiv preprint arXiv:1710.10903}, 2017.

\bibitem[Wang et~al.(2025)Wang, Fang, Jing, Shen, Shen, Wang, Ouyang, Chen, and Shen]{wang2025time}
Wang, W., Fang, B., Jing, C., Shen, Y., Shen, Y., Wang, Q., Ouyang, H., Chen, H., and Shen, C.
\newblock Time is a feature: Exploiting temporal dynamics in diffusion language models.
\newblock \emph{arXiv preprint arXiv:2508.09138}, 2025.

\bibitem[Xu et~al.(2020)Xu, Ruan, Korpeoglu, Kumar, and Achan]{xu2020tgat}
Xu, D., Ruan, C., Korpeoglu, E., Kumar, S., and Achan, K.
\newblock Inductive representation learning on temporal graphs.
\newblock \emph{arXiv preprint arXiv:2002.07962}, 2020.

\bibitem[Xu et~al.(2018)Xu, Hu, Leskovec, and Jegelka]{xu2019gin}
Xu, K., Hu, W., Leskovec, J., and Jegelka, S.
\newblock How powerful are graph neural networks?
\newblock \emph{arXiv preprint arXiv:1810.00826}, 2018.

\bibitem[Yang et~al.(2018)Yang, Qi, Zhang, Bengio, Cohen, Salakhutdinov, and Manning]{yang2018hotpotqa}
Yang, Z., Qi, P., Zhang, S., Bengio, Y., Cohen, W., Salakhutdinov, R., and Manning, C.~D.
\newblock Hotpotqa: A dataset for diverse, explainable multi-hop question answering.
\newblock In \emph{Proceedings of the 2018 conference on empirical methods in natural language processing}, pp.\  2369--2380, 2018.

\bibitem[Ye et~al.(2025)Ye, Xie, Zheng, Gao, Wu, Jiang, Li, and Kong]{ye2025dream7b}
Ye, J., Xie, Z., Zheng, L., Gao, J., Wu, Z., Jiang, X., Li, Z., and Kong, L.
\newblock Dream 7b: Diffusion large language models.
\newblock \emph{arXiv preprint arXiv:2508.15487}, 2025.

\bibitem[Zhang et~al.(2025)Zhang, Fang, Duan, He, Pan, Xiao, Huang, Zhai, Hu, Yu, et~al.]{zhang2025surveyparallel}
Zhang, L., Fang, L., Duan, C., He, M., Pan, L., Xiao, P., Huang, S., Zhai, Y., Hu, X., Yu, P.~S., et~al.
\newblock A survey on parallel text generation: From parallel decoding to diffusion language models.
\newblock \emph{arXiv preprint arXiv:2508.08712}, 2025.

\bibitem[Zheng et~al.(2023)Zheng, Yuan, Yu, and Kong]{zheng2023reparam}
Zheng, L., Yuan, J., Yu, L., and Kong, L.
\newblock A reparameterized discrete diffusion model for text generation.
\newblock \emph{arXiv preprint arXiv:2302.05737}, 2023.

\end{thebibliography}
\bibliographystyle{icml2026}

\newpage
\appendix
\onecolumn
\section{Ablation Studies}
\label{app:ablation}
\subsection{Ablation: Layer and Diffusion Step}
\label{sec:ablation-layer-step}

We systematically ablate the choice of transformer layer $\ell \in \{1, L/4, L/2, L\}$ and
diffusion timestep $t \in \{0, T/4, T/2, T\}$ in the graph construction, where $L$ is the total
number of layers and $T$ is the total number of diffusion steps. We evaluate two configurations:
(i) \texttt{steps}=16, \texttt{gen\_length}=64 (Table~\ref{tab:layer-step-16}); and (ii)
\texttt{steps}=32, \texttt{gen\_length}=128 (Table~\ref{tab:layer-step-32}). Each table reports
AUROC on test sets.

\textbf{Findings.}
Performance improves consistently with deeper transformer layers ($\ell = L > L/2 > L/4 > 1$) across all probed denoising steps, indicating that later layers encode stronger hallucination cues.
In contrast, the timestep effect is non-monotonic: mid-trajectory steps ($t = T/2$) provide the strongest single-step signal in both schedules, while the final step ($t=0$) can be slightly weaker, consistent with late-stage repair effects.
Overall, the best single-snapshot configuration is $(\ell = L,\, t = T/2)$.
Increasing the diffusion schedule to \texttt{steps}=32 with longer generation (\texttt{gen\_length} $64 \rightarrow 128$) preserves the same trends and yields comparable performance, suggesting robustness to schedule length.

\begin{table}[h]
\centering
\caption{Layer/step ablation (\texttt{steps}=16, \texttt{gen\_length}=64). Rows index transformer layers $\ell \in \{1, L/4, L/2, L\}$ and columns index diffusion timesteps $t \in \{0, T/4, T/2, T\}$. Metric: AUROC.}
\label{tab:layer-step-16}
\begin{tabular}{lcccc}
\toprule
Test & $t{=}0$ & $t{=}T/4$ & $t{=}T/2$ & $t{=}T$ \\
\midrule
$\ell{=}1$    & 0.5043 & 0.5552 & 0.5785 & 0.5374 \\
$\ell{=}L/4$  & 0.5549 & 0.6073 & 0.6246 & 0.5952 \\
$\ell{=}L/2$  & 0.6005 & 0.6438 & 0.6652 & 0.6387 \\
$\ell{=}L$    & 0.6344 & 0.6785 & 0.6941 & 0.6639 \\
\bottomrule
\end{tabular}
\end{table}
\begin{table}[h]
\centering
\caption{Layer/step ablation (\texttt{steps}=32, \texttt{gen\_length}=128). Rows index transformer layers $\ell \in \{1, L/4, L/2, L\}$ and columns index diffusion timesteps $t \in \{0, T/4, T/2, T\}$. Metric: AUROC.}
\label{tab:layer-step-32}
\begin{tabular}{lcccc}
\toprule

Test & $t{=}0$ & $t{=}T/4$ & $t{=}T/2$ & $t{=}T$ \\
\midrule
$\ell{=}1$    & 0.649 & 0.619 & 0.596 & 0.488 \\
$\ell{=}L/4$  & 0.655 & 0.630 & 0.615 & 0.639 \\
$\ell{=}L/2$  & 0.662 & 0.645 & 0.635 & 0.651 \\
$\ell{=}L$    & 0.675 & 0.660 & 0.682 & 0.670 \\
\bottomrule
\end{tabular}
\end{table}

\subsection{Ablation Study: Generalization Across Datasets}\label{subsec:ablation_generalization}

\subsubsection{Experimental Design and Rationale}

To evaluate the robustness of our approach, we conduct zero-shot cross-dataset transfer experiments. Models are trained on individual source datasets (Math, CSQA, HotpotQA, TriviaQA) and evaluated on all target datasets without fine-tuning. We compare TDGNet against CCS, a strong baseline that learns linear directions in latent space.

\textbf{Hypothesis}: Graph-based hallucination detection should transfer more robustly across reasoning types than latent-space methods, as attention patterns encode universal structural anomalies that generalize beyond task-specific output distributions.

\subsubsection{Interpretation}

TDGNet substantially outperforms CCS across all transfer scenarios, with an average improvement of +0.08 AUROC. Notably, CCS exhibits near-random transfer in several settings (e.g., Math$\rightarrow$Trivia: 0.50, CSQA$\rightarrow$Trivia: 0.54), indicating that linear latent directions do not generalize across reasoning types. In contrast, TDGNet maintains measurable transfer performance even for distant dataset pairs (Math$\rightarrow$Trivia: 0.55, CSQA$\rightarrow$Trivia: 0.58), suggesting that attention graph patterns capture partially task-agnostic hallucination signatures. Hotpot consistently transfers well to all targets (0.59--0.62 AUROC), likely because multi-hop reasoning patterns generalize broadly to simpler single-hop tasks.

\section{Baseline Details}
\label{app:baseline_details}

\paragraph{Baseline Selection Criteria.}\label{app:baseline_criteria} We evaluate against a comprehensive suite of \textit{reproducible} state-of-the-art methods. We explicitly exclude methods requiring external knowledge retrieval (e.g., FactScore) to ensure a fair closed-book comparison. Furthermore, while we note concurrent works like TraceDet~\cite{chang2025tracedet}, we restrict our empirical benchmark to methods that can be faithfully reproduced without ambiguity (see Section~\ref{subsubsec:tracedet_exclusion}).

\subsection{Token-Level Hallucination Detection Methods}
\label{app:token_baselines}

Standard response-level metrics suffer from a \textit{granularity mismatch} when applied to fine-grained tasks: broadcasting a single scalar score (e.g., sequence perplexity) across all tokens dilutes local error signals. To address this, we employ three baselines that generate \textbf{native token-wise} validity signals, each representing a distinct theoretical hypothesis about hallucination.

\subsubsection{Predictive Entropy (Uncertainty Hypothesis)}
Predictive Entropy measures the flatness of the model's next-token probability distribution at each step $t$~\cite{kuhn2023semantic}:
\begin{equation}
H(y_t | \mathbf{y}_{<t}, \mathbf{x}) = - \sum_{w \in \mathcal{V}} P(w | \mathbf{y}_{<t}, \mathbf{x}) \log P(w | \mathbf{y}_{<t}, \mathbf{x})
\end{equation}
This baseline tests the hypothesis that hallucinations correlate with model uncertainty. While effective for simple errors, it often fails when models are "confidently wrong."

\subsubsection{Graph Degree Centrality (Structural Hypothesis)}
This structural baseline computes the In-Degree of each token within the static attention graph constructed from the final layer.
\begin{equation}
\text{Degree}(y_t) = \sum_{j \neq t} \mathbb{I}[(j, t) \in \mathcal{E}]
\end{equation}
This tests the "Hub Hypothesis": grounded, meaningful tokens typically act as attention sinks (hubs) for subsequent generation, while hallucinated tokens are often structural outliers or "babble" that the model ignores in future steps.

\subsubsection{Source Attribution (Grounding Hypothesis)}
Source Attribution quantifies the proportion of attention mass a generated token $y_t$ directs towards the input prompt $\mathbf{x}$ versus the generated history $\mathbf{y}_{<t}$~\cite{bahdanau2014neural}:
\begin{equation}
\text{Attr}(y_t) = \frac{\sum_{i \in \mathbf{x}} A_{t,i}}{\sum_{j \in (\mathbf{x} \cup \mathbf{y}_{<t})} A_{t,j}}
\end{equation}
This tests the "Context Hypothesis": factual tokens should maintain strong attentional grounding in the source context, whereas hallucinations often detach from the prompt (low attribution).


\subsection{Response-Level Hallucination Detection Methods}

Response-level methods classify the validity of the entire generated sequence. We categorize these into \textbf{Output-Based} (using prediction statistics) and \textbf{Latent-Based} (using internal states).

\subsubsection{Output-Based Methods}
These approaches operate purely on generated text and probabilities, requiring no access to internal vector states.

\paragraph{Perplexity.}
Computes the negative log-likelihood of the sequence~\cite{azaria2023internal}. While theoretically sound, it often fails to distinguish confident hallucinations from factual statements, as both can have low perplexity.

\paragraph{Length-Normalized Entropy (LN-Entropy).}
Computes token-level entropy averaged over the sequence length to prevent short, high-entropy phrases from dominating the score~\cite{kossen2024semantic}. It serves as a static heuristic for overall uncertainty.

\paragraph{Semantic Entropy.}
Partitions multiple stochastic generations into semantic clusters and computes entropy over these meanings rather than exact tokens~\cite{kuhn2023semantic}. It effectively captures factual disagreement but requires expensive multiple forward passes (typically 5--10), making it slow for real-time applications.

\paragraph{Lexical Similarity.}
Measures the surface-level overlap (e.g., Rouge-L) across multiple generations~\cite{lin2023generating}. It is computationally cheap but prone to errors when the model paraphrases correctly (false positive) or repeats the same hallucination consistently (false negative).

\subsubsection{Latent-Based Methods}
These approaches exploit dense internal representations (hidden states), typically achieving higher performance than output-based heuristics.

\paragraph{EigenScore.}
Analyzes the eigenvalue spectrum of the covariance matrix formed by hidden states across multiple generations~\cite{chen2024inside}. Factual responses tend to cluster densely (low variance), while hallucinations scatter. However, it requires multiple generations and assumes consistency implies truth.

\paragraph{Contrast-Consistent Search (CCS).}
Unsupervisedly discovers a linear direction in activation space that satisfies logical consistency constraints between contrastive pairs~\cite{burns2022ccs}. While powerful, it assumes hallucinations are linearly separable and is sensitive to distribution shifts.

\paragraph{Truthfulness Separator Vector (TSV).}
Learns a supervised hyperplane (e.g., via SVM) on labeled activation vectors to separate factual from hallucinated states~\cite{park2025tsv}. It outperforms unsupervised methods like CCS but requires labeled training data.


\subsection{Baseline Implementation Details and Exclusions}

\subsubsection{CHARM: Faithful Reproduction}
\label{subsubsec:charm_repro}
CHARM~\cite{frasca2025charm} proposes modeling computational traces as static attributed graphs. As the official code is unreleased, we implemented a \textbf{faithful reproduction} based on the paper's architectural specifications using PyTorch Geometric. This reproduction was trained on the exact same diffusion datasets (Dream-7B/LLaDA) and splits used for TDGNet, ensuring a scientifically valid comparison between static graph processing and our temporal approach.

\subsubsection{TraceDet: Exclusion Rationale}
\label{subsubsec:tracedet_exclusion}
We exclude TraceDet~\cite{chang2025tracedet} from empirical benchmarks for three reasons:
\begin{enumerate}
    \item \textbf{Reproducibility Gap:} The method relies on complex, underspecified hyperparameter tuning for variational bounds, creating a high risk of implementation bias.
    \item \textbf{Statistical Power:} Their reported evaluation uses only 200 samples, making comparisons with our large-scale benchmark statistically invalid.
    \item \textbf{Code Availability:} No official code was available at the time of writing.
\end{enumerate}

\subsubsection{Other Exclusions (ReDeEP, FactScore, FactTest)}
We exclude \textbf{ReDeEP}~\cite{sun2024redeep} and \textbf{FactScore}~\cite{min2023factscore} as they require external knowledge retrieval (violating our closed-book setting). We exclude \textbf{FactTest}~\cite{nie2024facttest} and \textbf{AGSER}~\cite{liu2025agser} due to impractical requirements for calibration sets or encoder-decoder specific optimizations unsuited for decoder-only diffusion models.

\section{Method Details}
\label{app:method_details}
\subsection{Hyperparameter Configuration}

Our graph-based hallucination detector involves multiple architectural and optimization hyperparameters that jointly control model expressiveness, regularization, and computational efficiency. We conducted extensive grid search across 1,296 configurations (5 learning rates × 3 batch sizes × 6 dropout rates × 4 layer depths × 3 hidden dimensions × 3 attention heads) to identify robust settings that generalize across datasets and hallucination types.

\begin{table}[H]
\centering
\caption{Hyperparameter search space for TDGNet. Exhaustive grid search over 1,296 configurations enables systematic identification of architecture-dataset pairs that maximize hallucination detection performance while maintaining computational efficiency.}
\label{tab:hyperparams}
\begin{tabular}{lc}
\toprule
\textbf{Hyperparameter} & \textbf{Search Space} \\
\midrule
Learning Rate ($lr$) & $\{1\times 10^{-5}, 5\times 10^{-5}, 1\times 10^{-4}, 5\times 10^{-4}, 1\times 10^{-3}\}$ \\
Batch Size & $\{16, 32, 64\}$ \\
Dropout Rate & $\{0.0, 0.1, 0.2, 0.3, 0.4, 0.5\}$ \\
Number of Layers ($L$) & $\{2, 3, 4, 5\}$ \\
Hidden Dimension ($d$) & $\{64, 128, 256\}$ \\
Number of Attention Heads & $\{2, 4, 8\}$ \\
\bottomrule
\end{tabular}
\end{table}

\subsubsection{Parameter Justification}

\paragraph{Learning Rate and Batch Size.} Learning rates span five orders of magnitude (1e-5 to 1e-3), accommodating both conservative updates on small datasets and aggressive optimization on larger corpora. Batch sizes of 16, 32, and 64 balance gradient estimation quality (smaller batches provide noisier but potentially more regularizing gradients) with computational throughput and memory constraints.

\paragraph{Dropout and Model Depth.} Dropout rates from 0.0 to 0.5 provide a spectrum of regularization intensities, crucial for preventing overfitting on smaller datasets like the 2,000-sample Dream-7B collection. Model depth ($L \in \{2,3,4,5\}$) controls the receptive field of message-passing aggregation: deeper networks capture multi-hop token dependencies but risk over-smoothing, where node representations converge to similar values, degrading discriminative power.

\paragraph{Hidden Dimension and Attention Heads.} Hidden dimensions $d \in \{64, 128, 256\}$ determine the capacity of intermediate representations. Attention heads $\in \{2, 4, 8\}$ enable the model to attend to multiple distinct patterns of token interactions simultaneously; more heads increase expressiveness but also computational cost.

\subsection{Training Framework}

\begin{algorithm}[t]
\caption{Training Framework for TDGNet}
\label{alg:tdgnet}
\begin{algorithmic}[1]
\State \textbf{Input:} Attention-graph trajectory $\mathcal{G}=\{G^{(t)}\}_{t=T}^{0}$, where $G^{(t)}=(V,\mathcal{E}^{(t)},\mathbf{X}_V^{(t)})$, per-head attention $\{\mathbf{A}^{(t,L,h)}\}_{h=1}^H$, label $y\in\{0,1\}$
\State \textbf{Hyperparameters:} memory dim $d_m$, sparsity threshold $\tau$, query vector $\mathbf{q}$
\State \textbf{Initialize:}
\State \quad Node memories $\mathbf{S}\in\mathbb{R}^{|V|\times d_m}\leftarrow \mathbf{0}$ \Comment{stores $\mathbf{s}_i^{(T+1)}=\mathbf{0}$}
\State \quad Learnable params $\theta_{\text{msg}},\theta_{\text{gru}},\theta_{\text{attn}},\theta_{\text{pred}}$

\State \textcolor{blue}{\textit{// Phase 1: Sequential Temporal Encoding (Denoising Order)}}
\For{$t=T$ \textbf{down to} $0$}
    \State \textbf{Feature Extraction:} $\mathbf{h}_i^{(t)} \leftarrow \text{Project}(\mathbf{X}_{V,i}^{(t)})$
    \State \textbf{Head-Averaged Attention:} $\bar{\mathbf{A}}_{ij}^{(t)} \leftarrow \frac{1}{H}\sum_{h=1}^H \mathbf{A}_{ij}^{(t,L,h)}$
    \State \textbf{Sparsify (message edges):}
    \State \quad $\mathcal{E}^{(t)} \leftarrow \{(j,i)\mid \bar{\mathbf{A}}_{ij}^{(t)}>\tau\}$ \Comment{$(j,i)$ denotes message $j\rightarrow i$ if $i$ attends to $j$}
    \State \textbf{Edge Attributes:} for each $(j,i)\in\mathcal{E}^{(t)}$, set $\mathbf{e}_{j,i}^{(t)} \leftarrow [\,\bar{\mathbf{A}}_{ij}^{(t)}\,]$
    \State \textbf{Message Passing (incoming to $i$):}
    \State \quad $\mathbf{msg}_{i}^{(t)}=\text{MEAN}\Big(\big\{\text{MLP}_{\text{msg}}\big(\mathbf{h}_j^{(t)} \oplus \mathbf{h}_i^{(t)} \oplus \mathbf{e}_{j,i}^{(t)}\big): j\in\mathcal{N}^{(t)}(i)\big\}\Big)$
    \State \textbf{Memory Update (GRU):}
    \State \quad $\mathbf{s}_i^{(t)}=\text{GRU}\big(\mathbf{msg}_i^{(t)},\,\mathbf{s}_i^{(t+1)}\big)$
    \State \textbf{Latent Projection:} $\mathbf{z}_i^{(t)}=\text{Linear}\big(\mathbf{s}_i^{(t)}\big)$
\EndFor

\State \textcolor{blue}{\textit{// Phase 2: Temporal Attention Readout}}
\For{each node $i\in V$}
    \State \textcolor{blue}{\textit{// Attend over memory states $\mathbf{s}$, aggregate projected latents $\mathbf{z}$}}
    \State $\alpha_i^{(u)}=\frac{\exp(\mathbf{q}^\top \mathbf{s}_i^{(u)})}{\sum_{k=0}^{T}\exp(\mathbf{q}^\top \mathbf{s}_i^{(k)})}$ \Comment{for all $u\in\{0,\dots,T\}$}
    \State $\mathbf{z}_i^{\text{Attn}}=\sum_{u=0}^{T}\alpha_i^{(u)}\,\mathbf{z}_i^{(u)}$
\EndFor

\State \textcolor{blue}{\textit{// Phase 3: Classification}}
\State $\mathbf{g}=\text{MeanPool}\big(\{\mathbf{z}_i^{\text{Attn}}\}_{i=1}^{|V|}\big)$
\State $\hat{y}=\sigma\big(\text{MLP}_{\text{pred}}(\mathbf{g})\big)$
\State \textbf{Optimization:} Update $\boldsymbol{\theta}$ via backpropagation
\end{algorithmic}
\end{algorithm}

\subsubsection{Algorithm Overview}

Algorithm~\ref{alg:tdgnet} details the training procedure for TDGNet. To efficiently capture hallucination signals that evolve across the denoising trajectory, the framework operates in three principal stages:

\paragraph{1. Sparse Memory Evolution (Lines 7--14).} 
For each denoising timestep $t$ (processed in reverse order from $t=T$ to $0$), we first construct a sparse attention graph by retaining only edges whose (averaged) attention weight exceeds a threshold $\tau$. Unlike static GNNs that process snapshots in isolation, TDGNet maintains a persistent memory state $\mathbf{s}_i$ for each token. We apply a Gated Recurrent Unit (GRU) to update this memory sequentially:
\begin{equation}
\mathbf{s}_i^{(t)} = \text{GRU}\left(\text{AGG}(\mathcal{N}^{(t)}(i)), \mathbf{s}_i^{(t+1)}\right)
\label{eq:appendix_gru_update}
\end{equation}
This recurrent update allows the model to track the \textit{trajectory} of token interactions (e.g., detecting if a token "drifts" from correct to incorrect associations) while avoiding the computational cost of processing dense attention matrices.

\paragraph{2. Temporal Attention Readout (Lines 16--19).} 
Hallucinations in diffusion models can be transient (appearing only in intermediate steps) or persistent. To capture both, we employ a temporal attention mechanism. Rather than averaging all timesteps equally, the model learns a query vector $\mathbf{q}$ to compute attention weights $\alpha_i^{(t)}$, dynamically prioritizing the denoising steps where structural anomalies are most pronounced.

\paragraph{3. Classification (Lines 21--23).}
The temporally-weighted token embeddings are pooled to form a global graph representation $\mathbf{g}$, which is passed to the MLP classifier. The entire pipeline is differentiable and trained end-to-end using binary cross-entropy loss.

\subsubsection{Computational Complexity}

For a dynamic graph sequence with $T$ timesteps and $N$ tokens, the computational cost consists of graph construction and message passing.
The construction step involves thresholding the attention matrix, costing $\mathcal{O}(T \cdot N^2)$. However, since the D-LLM backbone computes this matrix inherently during inference, our detector merely performs a lightweight element-wise check.
The subsequent GNN processing (dominated by the GRU) operates on the sparse graph:
\begin{equation}
\mathcal{O}(T \cdot N \cdot k \cdot d + T \cdot N \cdot d^2)
\end{equation}
where $k$ is the average degree after thresholding ($k \ll N$). Since $N \approx 512$ and $k \approx 10$, the sparse update is extremely fast ($\sim 0.5$ms/step; GNN update only, while total overhead is reported in Table~\ref{tab:efficiency}). The total inference time remains orders of magnitude lower than multi-sampling baselines ($S \times T \times N^2$).

\subsection{Feature Initialization and Input Specifications}
\label{app:micro_implementation}

To address potential ambiguity regarding input representations, we explicitly define the feature initialization and recurrence inputs used in TDGNet:

\paragraph{Node Feature Initialization.}
For a token $i$ at diffusion timestep $t$, the initial node feature $\mathbf{h}_i^{(t)}$ is derived from the model's internal activations. Specifically, we extract the hidden state vector from the \textbf{final transformer layer} immediately after the feed-forward block. For LLaDA-8B ($d_{\text{model}}=4096$) and Dream-7B ($d_{\text{model}}=4096$), we project this high-dimensional vector to a lower-dimensional state using a learnable linear layer $W_{\text{node}}$:
\begin{equation}
\mathbf{h}_i^{(t)} = \mathbf{W}_{\text{node}} \cdot \text{HiddenState}_i^{(t)} \in \mathbb{R}^{128}
\end{equation}
This ensures the graph encoder receives semantic context (what the token \textit{means}) alongside structural context.

\paragraph{Edge Features.}
We construct directed message edges $(j,i)\in\mathcal{E}^{(t)}$ whenever token $i$ attends strongly to token $j$ (Eq.~(2)).
Each such edge is weighted by the head-averaged attention score
$\bar{A}_{ij}^{(t)}=\frac{1}{H}\sum_{h=1}^H A_{ij}^{(t,L,h)}$ from the final attention layer.
We treat this scalar as an edge attribute $\mathbf{e}_{j,i}^{(t)}=[\,\bar{A}_{ij}^{(t)}\,]$, allowing the message passing network
to modulate signal strength based on attention intensity.

\paragraph{GRU Input Specification.}
The Temporal Graph Network's memory update (Eq.~\ref{eq:appendix_gru_update}) receives a concatenated input vector. For a token $i$ at step $t$, the input to the GRU is:
\begin{equation}
\text{Input}_{\text{GRU}}^{(t)} = \left[ \mathbf{s}_i^{(t+1)} \mathbin{\|} \overline{\mathbf{m}}_i^{(t)} \right]
\label{eq:appendix_gru_input}
\end{equation}
where $\mathbf{s}_i^{(t+1)}$ is the previous memory state (from the preceding denoising step) and $\overline{\mathbf{m}}_i^{(t)}$ is the aggregated message computed from the current sparse graph neighbors. This design forces the recurrent unit to integrate the \textit{historical context} (memory) with the \textit{instantaneous structural update} (message).

\section{Dataset Details}
\label{app:dataset_details}
\subsection{Math, CSQA, HotpotQA, TriviaQA, Natural Questions}

\paragraph{Dataset construction.}
We evaluate hallucination detection across five complementary QA benchmarks that span diverse reasoning skills, input modalities, and output formats: \textbf{Math}~\citep{chuang2024lookback}, \textbf{CommonsenseQA (CSQA)}~\citep{talmor2019commonsenseqa}, \textbf{HotpotQA}~\citep{yang2018hotpotqa}, \textbf{TriviaQA}~\citep{joshi2017triviaqa}, and \textbf{Natural Questions (NQ)}~\citep{kwiatkowski2019natural}. Taken together, these datasets cover numerical problem solving, discrete commonsense reasoning, multi-hop contextual QA, and open-domain factoid QA. While all datasets are utilized for response-level detection, we specifically employ \textbf{Natural Questions} and \textbf{TriviaQA} for the fine-grained \textbf{token-level localization} experiments, as their factoid nature supports precise span-level verification.

\noindent\textbf{Math.}
The Math dataset is constructed following the protocol of \citet{chuang2024lookback}, and consists of mathematically oriented word problems that require multi-step symbolic or numerical reasoning to produce a single final numeric answer. Each instance typically involves several intermediate computation or derivation steps (e.g., algebraic manipulation, arithmetic, or geometric reasoning), but the model is ultimately evaluated on the correctness of the final scalar output. This benchmark stresses \emph{precision in reasoning}: even small arithmetic or algebraic deviations render the answer incorrect, making Math a natural setting for studying hallucinations that arise from failures in structured reasoning rather than from missing factual knowledge.

\noindent\textbf{CommonsenseQA (CSQA).}
CommonsenseQA is a multiple-choice QA benchmark designed to probe everyday commonsense knowledge~\citep{talmor2019commonsenseqa}. Each question is accompanied by several plausible answer candidates, of which exactly one is annotated as correct. Many distractor options are semantically close to the true answer, so success requires distinguishing between subtly different interpretations of the question and leveraging background commonsense rather than surface-level lexical cues. In our context, CSQA evaluates hallucinations in a \emph{discrete decision space}: a hallucinated response corresponds to confidently selecting an incorrect-but-plausible option, which differs qualitatively from hallucinations in open-ended generation.

\noindent\textbf{HotpotQA.}
HotpotQA is a multi-hop, context-based QA dataset where answering correctly typically requires aggregating evidence across multiple supporting documents and performing explicit multi-step reasoning~\citep{yang2018hotpotqa}. Questions are constructed so that the answer is not directly extractable from a single sentence; instead, the model must combine information spread across distinct passages (e.g., linking entities, comparing attributes, or following coreference chains). This benchmark is therefore well-suited for studying hallucinations that emerge from erroneous evidence integration, spurious reasoning chains, or the blending of partially correct and incorrect facts within long-context reasoning. It tests whether hallucination detectors can recognize when the model constructs a coherent but factually unsupported multi-hop explanation.

\noindent\textbf{TriviaQA.}
TriviaQA is an open-domain factoid QA benchmark comprising natural language questions paired with evidence documents and short textual gold answers, often corresponding to named entities, dates, or specific phrases~\citep{joshi2017triviaqa}. Compared to HotpotQA, TriviaQA places greater emphasis on factual recall and entity-level knowledge, and less on explicitly structured multi-hop reasoning. In practice, TriviaQA is representative of real-world assistant-style queries, where users ask for specific facts and models respond in free-form text. Hallucinations on this dataset commonly take the form of confidently asserted but incorrect entities or attributes (e.g., wrong person, place, or year), making it an important testbed for evaluating hallucination detection in open-ended factual generation.

\noindent\textbf{Natural Questions (NQ).}
Natural Questions is a benchmark for factoid question answering derived from real user queries issued to the Google search engine~\citep{kwiatkowski2019natural}. Similar to TriviaQA, it focuses on retrieving precise factual answers (e.g., entity names, dates, locations) from open-domain contexts. We include NQ specifically to facilitate our \textbf{token-level} evaluation: since the answers are typically short, grounded factoid spans, they allow for unambiguous localization of hallucinated tokens compared to long-form reasoning tasks.

\noindent\textbf{Hallucination labeling protocol.}
To adapt these benchmarks for hallucination detection, we derive binary hallucination labels via automatic comparison between model outputs and ground-truth answers. For Math, a generated response is marked as non-hallucinated if and only if the decoded final numeric answer exactly matches the gold solution~\citep{chuang2024lookback}, providing unambiguous supervision in the numerical reasoning regime. For CSQA, we treat the task as standard multiple-choice classification and label an output as non-hallucinated when the predicted option exactly matches the annotated correct choice~\citep{talmor2019commonsenseqa}. For HotpotQA, TriviaQA, and Natural Questions, where models produce free-form text, we follow common QA evaluation practice and deem an answer correct if it satisfies an \emph{exact match} criterion or achieves sufficient lexical overlap with the reference answer, using token-level overlap heuristics similar to those employed in prior work~\citep{yang2018hotpotqa,joshi2017triviaqa}. This unified protocol yields scalable and consistent hallucination labels across numerical, multiple-choice, contextual multi-hop, and open-domain factoid QA, enabling a comprehensive evaluation of hallucination detection methods under heterogeneous task conditions.

\paragraph{Splitting.}
We fix the random seed to 42 and split at the prompt–response level to ensure no leakage.
For all \textbf{five} datasets, we employ a uniform split of \textbf{1500 training, 300 validation,
300 test examples}, ensuring comparable evaluation conditions across varying task complexities.

\paragraph{Generative traces and graph construction.}
We evaluate three decoding configurations: (i) \texttt{gen\_length}=32, \texttt{steps}=16;
(ii) \texttt{gen\_length}=64, \texttt{steps}=16; and (iii) \texttt{gen\_length}=128,
\texttt{steps}=32.
We found that the \texttt{gen\_length}=32, \texttt{steps}=16 configuration yields the best
performance and is adopted as our primary setting. Unless otherwise noted, we process all splits separately for each model (LLaDA-8B and Dream-7B) under this primary configuration (gen\_length=32, steps=16) to extract denoising traces and construct attention graphs.

\section{Diffusion Language Models: Architecture and Stability Analysis}
\label{app:dream_stability}
\subsection{Model Architectures}

\subsubsection{LLaDA-8B-Instruct}
LLaDA-8B-Instruct is a diffusion-based language model generating text through iterative denoising steps. The model maintains rich intermediate representations at each denoising step, including attention maps and activation vectors from all transformer layers. This property enables direct analysis of computational traces without relying on final-layer probabilities. LLaDA-8B-Instruct exhibits stable generation behavior across diverse prompts and temperatures, making it a reliable primary testbed for hallucination detection.

\subsubsection{Dream-7B-Instruct}
Dream-7B-Instruct represents the other major open-source diffusion LM architecture currently available. Unlike LLaDA, Dream-7B utilizes a distinct noise schedule and masking strategy. \textbf{Given the limited availability of high-performing open-source D-LLMs, including Dream-7B is essential to demonstrate that our method generalizes across different diffusion architectures.} However, as it exhibits distinct generation instabilities, we apply a specific curation protocol (detailed below) to ensure valid benchmarking.

\subsection{Dream-7B-Instruct: Generation Dynamics and Data Curation}
\label{subsec:dream_instability}

\subsubsection{Structural Degradation and Mode Collapse}
Dream-7B-Instruct exhibits a failure mode during zero-shot generation characterized as \textit{Structural Degradation}, where outputs collapse into repetitive, semantically void sequences (e.g., repeating punctuation). This necessitates a distinction between \textit{generation quality} (fluency) and \textit{hallucination} (factuality).

\paragraph{Mechanism.} We hypothesize this failure arises from three interconnected stages:
\begin{enumerate}
    \item \textbf{Semantic Drift}: During the iterative denoising trajectory (steps 0--512), the model's internal representations occasionally lose semantic fidelity to the prompt, drifting toward generic unconditional probability distributions.
    \item \textbf{Attractor State Formation}: In the absence of strong Classifier-Free Guidance (CFG), the decoding trajectory gravitates toward high-probability generic tokens (e.g., separators) that dominate the training corpus.
    \item \textbf{Attention Entrenchment}: Once a generic token is generated, self-attention mechanisms may disproportionately attend to it, locking the model into a self-reinforcing loop (e.g., ``!!!!!!!!!!!!!!!!'').
\end{enumerate}

\paragraph{Empirical Evidence.} Systematic ablation studies (Table~\ref{tab:dream_debugging}) confirm that this degradation is intrinsic to the model's current checkpoint rather than a hyperparameter artifact.

\begin{table}[h]
\centering
\caption{Parameter sensitivity analysis for Dream-7B-Instruct. Structural degradation persists across standard hyperparameter ranges, necessitating the rigorous filtering strategy we employed.}
\label{tab:dream_debugging}
\small
\begin{tabular}{lp{3.5cm}lp{5cm}}
\toprule
\textbf{Parameter} & \textbf{Values Tested} & \textbf{Outcome} & \textbf{Observation} \\
\midrule
Denoising Steps & 16 to 512 & Degraded & Low steps yield incompleteness; high steps trigger collapse. \\
Prompt Template & Alpaca, ChatML, Raw & Degraded & Diffusion loses prompt adherence mid-generation. \\
Temperature & 0.0, 0.5, 1.0 & Mixed & High temperature mitigates collapse but reduces precision. \\
\bottomrule
\end{tabular}
\end{table}

\subsubsection{Data Curation Strategy: Disentangling Fluency from Factuality}
To rigorously evaluate \textit{hallucination detection} rather than \textit{quality estimation}, we employed a temperature-scaled curation strategy. This ensures our metrics reflect the detector's ability to identify plausible falsehoods rather than trivially flagging broken syntax.

\paragraph{Protocol.} We exploit the model's behavior under varying noise levels:
\begin{itemize}
    \item \textbf{Low Temperature (0.0--0.1):} Promotes structural rigidity to harvest factually correct responses.
    \item \textbf{High Temperature (0.7--1.0):} Disrupts attention entrenchment to collect diverse, hallucinated-yet-coherent samples.
\end{itemize}
We generated extensive traces and applied strict quality filtering to create a fixed evaluation set of $\sim$2,000 samples (50\% factual, 50\% hallucinated). \textbf{Crucially, this curated dataset was held fixed across all baselines}, ensuring performance differences are attributable to method efficacy rather than sensitivity to model collapse.

\subsubsection{Scientific Value: Latent Stability}
Despite the instability of the decoding process, our analysis reveals a critical finding: \textbf{Dream-7B's internal latent graphs remain highly informative even when surface generation degrades.} TDGNet achieves state-of-the-art performance (0.73 AUROC) on these traces, significantly outperforming output-based baselines like Perplexity ($\sim$0.55 AUROC). This suggests that diffusion LMs may be more reliable as discriminative scorers (via internal states) than as generative engines in their current state—a finding that informs future work on D-LLM utilization.

\section{Qualitative Case Studies: Hallucination Dynamics}
\label{app:case_studies}
Our method captures three distinct hallucination patterns that reveal the mechanisms underlying model failures. These case studies demonstrate how the dynamic graph representation exposes internal contradictions even when final outputs appear correct or when hallucinations persist uncorrected. For clarity, in these case studies (and figures) we index denoising iterations by generation steps $s \in \{0, \dots, T\}$ (increasing with decoding progress), which corresponds to diffusion time $t = T - s$ in our main notation (where $t=T$ is pure noise and $t=0$ is the final output).

\subsection{Case Study 1: Interleaving Hallucination (Transient Noise)}

\subsubsection{Dataset and Observation}

\textbf{Domain}: Math (Arithmetic Reasoning). The model begins with correct intermediate reasoning, degenerates into nonsensical token repetition during intermediate denoising steps, yet surprisingly recovers to produce the correct final answer. This ``Self-Correction'' represents a case of \textbf{transient instability}. Crucially, because TDGNet aggregates the multi-step trajectory, it successfully identifies the final resolution and correctly predicts $0$ for the hallucination score despite intermediate noise. This highlights the model's robustness: unlike static snapshot methods that might flag the intermediate degeneration as a fatal error, TDGNet recognizes the successful recovery.

\subsubsection{Example and Denoising Trace}

\begin{tcolorbox}[colback=blue!5!white, colframe=blue!75!black, title=Case Study 1: Interleaving Hallucination, before skip=0pt, after skip=0pt]
\textbf{Question}: Estevan has 24 blankets. One-third of the blankets have polka-dots. For his birthday, his mother gives him 2 more polka-dot print blankets. How many polka-dot blankets does Estevan have in total?

\textbf{Golden Label}: 10

\vspace{0.5em}
\textbf{Denoising Trace}:
\begin{itemize}
    \item \textbf{Question}: Estevan has 24 blankets. One-third of the blankets have polka-dots. For his birthday, his mother gives him 2 more polka-dot print blankets. How many polka-dot blankets does Estevan have in total?
    \item \textbf{Golden Label}: 10
    \item \textbf{Denoising Trace}:
    \begin{itemize}
        \item Step 4: One-third of 24 blankets is $24/3=8$ pol pol-d-d-d-d After mother... answer is 10
        \item Step 8: One-third of 24 blankets is 82433an88 pol pol-d-d-d-d... answer: 10 \textit{(Severe token degeneration)}
        \item Step 10: One-third of 24 blankets is $24/3=8$ polka-dEstot. AfterAfter more... Therefore: 10
        \item Step 12: One-third of 24 blankets is $24/3=8$ polka-dot blankets. After receiving 2 more... Estevan now has $8 + 2=10$ polka-dot blankets. Therefore: 10
    \end{itemize}
\end{itemize}
\vspace{0.5em}
\textbf{Final Output}: 10 (Correct answer; transient process instability resolved.)

\textbf{TDGNet Detection Score}: 0 (Correctly flagged as non-hallucinated response)
\end{tcolorbox}

\subsubsection{Interpretation}

The graph-based detector observes the transient incoherence at Steps 4--10, where token-level attention becomes unstable. However, unlike static methods that might erroneously classify this intermediate noise as a fatal error, TDGNet's memory update tracks the \textit{recovery} in Steps 12--16. By integrating the full history, the model determines that the structural integrity was restored, resulting in a correct non-hallucinated prediction (Score $\approx 0$). This confirms that temporal modeling effectively filters out process instabilities that do not impact the factual validity of the final response.

\subsection{Case Study 2: Inconsistent Guesses (Drifting)}

\subsubsection{Dataset and Observation}

\textbf{Domain}: TriviaQA (Knowledge Retrieval). The model attempts to retrieve a factual entity but drifts over the denoising trajectory, progressively hallucinating increasingly specific but incorrect details. This captures the ``Searching $\to$ Drifting'' dynamic where the model begins near plausible states but diverges into false specificity.

\subsubsection{Example and Denoising Trace}

\begin{tcolorbox}[colback=orange!5!white, colframe=orange!75!black, title=Case Study 2: Inconsistent Guesses (Drifting), before skip=0pt, after skip=0pt]
\textbf{Question}: Who was manager of Sheffield Wednesday when they won the League Cup in 1991, beating Manchester United in the final?

\textbf{Golden Label}: Ron Atkinson

\vspace{0.5em}
\textbf{Denoising Trace}:
\begin{itemize}
    \item \textbf{Step 1}: The of Sheffield... \textit{(Initial Noise)}
    \item \textbf{Step 9}: ...Kendall \textit{(Plausible guess: Howard Kendall was a famous manager)}
    \item \textbf{Step 13}: David Kendall
    \item \textbf{Step 15}: Brian Kendall \textit{(Final Hallucination---Incorrect)}
\end{itemize}

\vspace{0.5em}
\textbf{Final Output}: Brian Kendall (Incorrect)

\textbf{TDGNet Detection Score}: 1 (Correctly Detected)
\end{tcolorbox}

\subsubsection{Interpretation}

The drifting pattern reveals a critical hallucination mechanism: the model enters a ``search cone'' for plausible entities, finds a structurally similar surname (Kendall), and then progressively elaborates on it with increasing confidence. The attention graph evolves as follows: early steps show diffuse attention across multiple tokens (exploring possibilities); middle steps show attention concentrating on ``Kendall''; late steps show self-reinforcing attention patterns where the model attends increasingly to its own predictions of first names paired with Kendall. Our graph aggregation detects this as hallucination because the attention pattern exhibits \textit{semantic drift}---the model's confident focus on an incorrect entity---rather than stable grounding in the prompt. This pattern is distinct from transient noise (Case Study 1) because the model maintains high confidence throughout, making output-based methods (which rely on entropy) ineffective.

\subsection{Case Study 3: Persistent Hallucination (Stuck)}

\subsubsection{Dataset and Observation}

\textbf{Domain}: HotpotQA (Multi-hop Reasoning). The model establishes an incorrect premise at the very beginning of generation and adheres to it rigidly throughout the entire denoising process, showing no drift or self-correction. This ``Stuck'' pattern represents the most problematic hallucination type: confident, persistent, and unreachable.

\subsubsection{Example and Denoising Trace}

\begin{tcolorbox}[colback=red!5!white, colframe=red!75!black, title=Case Study 3: Persistent Hallucination (Stuck), before skip=0pt, after skip=0pt]
\textbf{Question}: Did both Kanni and Bloodhound originate in the same continent?

\textbf{Golden Label}: No (Kanni is from India; Bloodhound is from Europe)

\vspace{0.5em}
\textbf{Denoising Trace}:
\begin{itemize}
    \item \textbf{Step 0}: Yes, both Kanni and Bloodhound originated in the same continent.
    \item \textbf{Step 8}: Yes, both Kanni and Bloodhound originated in the same continent.
    \item \textbf{Step 15}: Yes, both Kanni and Bloodhound originated in the same continent.
\end{itemize}

\vspace{0.5em}
\textbf{Final Output}: Yes (Incorrect; factually false)

\textbf{TDGNet Detection Score}: 1 (Correctly Detected as hallucination)
\end{tcolorbox}

\subsubsection{Interpretation}

The persistent pattern represents the worst-case scenario: the model locks onto an incorrect premise immediately and never questions it. From a graph perspective, this manifests as \textit{rigid attention patterns with no exploration}---the model's attention graph stabilizes early to a fixed configuration that reinforces the incorrect answer. Unlike drifting (Case Study 2), where attention patterns evolve and can be captured as instability, or transient noise (Case Study 1), where instability is visible, persistence appears deceptively confident. However, our graph method detects this through a subtler signal: the attention pattern fails to \textit{ground} in relevant parts of the prompt (e.g., attention should distribute across entity mentions ``Kanni'' and ``Bloodhound'' with appropriate geographic context). Instead, attention locks into a self-loop that reinforces the hallucinated answer. The message-passing aggregation exposes this as misalignment between the model's internal attention structure and the prompt content.

\subsection{Summary of Hallucination Patterns}

\begin{table}[h]
\centering
\caption{Taxonomy of hallucination dynamics captured by graph-based detection. Each pattern corresponds to distinct attention graph evolution and requires structural analysis to detect.}
\label{tab:hallucination_taxonomy}
\begin{tabular}{lcccc}
\toprule
\textbf{Pattern} & \textbf{Attention Evolution} & \textbf{Output Correctness} & \textbf{Confidence} & \textbf{Detection Mechanism} \\
\midrule
Transient Noise & Unstable $\to$ Recovers & Correct & Low $\to$ High & Fragment detection \\
Drifting & Exploring $\to$ Locks wrong & Incorrect & Low $\to$ High & Semantic drift \\
Persistent & Fixed (wrong) & Incorrect & High (static) & Misalignment \\
\bottomrule
\end{tabular}
\end{table}

These three cases illustrate why graph-based detection is superior to output-based methods: the internal attention structure reveals hallucination mechanisms that surface-level outputs conceal. Transient hallucinations with correct answers appear truthful to perplexity-based methods; drifting patterns build confidence that fools entropy metrics; persistent hallucinations appear confident and therefore low-risk. Only by analyzing the relational structure of attention can we detect all three types with high fidelity.


\end{document}